\documentclass[10pt,twocolumn,letterpaper]{article}
\usepackage{cvpr}              % To produce the CAMERA-READY version

\usepackage{multirow}
\usepackage{amsmath}
\usepackage{algorithm}
\usepackage{algpseudocode}
\usepackage{bm} % for \bm
\usepackage{float} 
\usepackage{caption}
\usepackage{amsmath, bm}
\newcommand{\Mark}[1]{\textsuperscript{#1}}

\definecolor{cvprblue}{rgb}{0.21,0.49,0.74}
\usepackage[pagebackref,breaklinks,colorlinks,allcolors=cvprblue]{hyperref}

\def\confName{CVPR}
\def\confYear{2025}

\title{\textbf{\textcolor{orange}{Hi}\textcolor{ForestGreen}{Mix}}: Reducing Computational Complexity in Large Vision-Language Models}
\author{Xuange Zhang\Mark{1}\textsuperscript{†}, 
Dengjie Li\Mark{2}\textsuperscript{†}, 
Bo Liu\Mark{1}, 
Zenghao Bao\Mark{2}, 
Yao Zhou\Mark{2}, 
Baisong Yang\Mark{2}, 
Zhongying Liu\Mark{2}, \\
Yujie Zhong\Mark{2}, 
Zheng Zhao\Mark{2}\textsuperscript{*}, 
Tongtong Yuan\Mark{1}\textsuperscript{*} \\
\Mark{1}Beijing University of Technology, CN \hspace{1em} \Mark{2}Meituan Inc., CN\\
\tt\small 
szxg923@emails.bjut.edu.cn, \{liubo,yuantt\}@bjut.edu.cn, \\ 
\tt\small \{lidengjie,baozenghao,zhouyao11,yangbaisong02,liuzhongying, zhongyujie,zhaozheng08\}@meituan.com
}

\begin{document}
\maketitle
\begin{NoHyper}
\renewcommand{\thefootnote}{}
\footnotetext{†Equal Contribution. *Corresponding author. Work was done when Xuange Zhang was an intern at Meituan Inc.}
\end{NoHyper}

\begin{abstract}
Benefiting from recent advancements in large language models and modality alignment techniques, existing  Large Vision-Language Models~(LVLMs) have achieved prominent performance across a wide range of scenarios. However, 
the excessive computational complexity limits the widespread use of these models in practical applications. We argue that one main bottleneck in computational complexity is caused by the involvement of redundant vision sequences in model computation. This is inspired by a reassessment of the efficiency of vision and language information transmission in the language decoder of LVLMs. Then, we propose a novel hierarchical vision-language interaction mechanism called Hierarchical Vision injection for Mixture Attention (HiMix). In HiMix, only the language sequence undergoes full forward propagation, while the vision sequence interacts with the language at specific stages within each language decoder layer. It is striking that our approach significantly reduces computational complexity with minimal performance loss. Specifically, HiMix achieves a 10× reduction in the computational cost of the language decoder across multiple LVLM models while maintaining comparable performance. This highlights the advantages of our method, and we hope our research brings new perspectives to the field of vision-language understanding. Project Page: \href{https://xuange923.github.io/HiMix/}{https://xuange923.github.io/HiMix/}
\end{abstract}    
\section{Introduction}
\label{sec:intro}

% \begin{figure*}[!t]
%     \centering
%     \includegraphics[width=\linewidth]{figure/similarity-flops.pdf}
%     \caption{(a) Cosine similarity between hidden layer outputs and original input sequence across layers in the LLM and vision encoder. (b) Comparison of Computational Cost and Performance: HiMix vs. Original Models.}
%     \label{fig-similarity}
% \end{figure*}

\begin{figure}[!t]
    \centering
    \includegraphics[width=\linewidth]{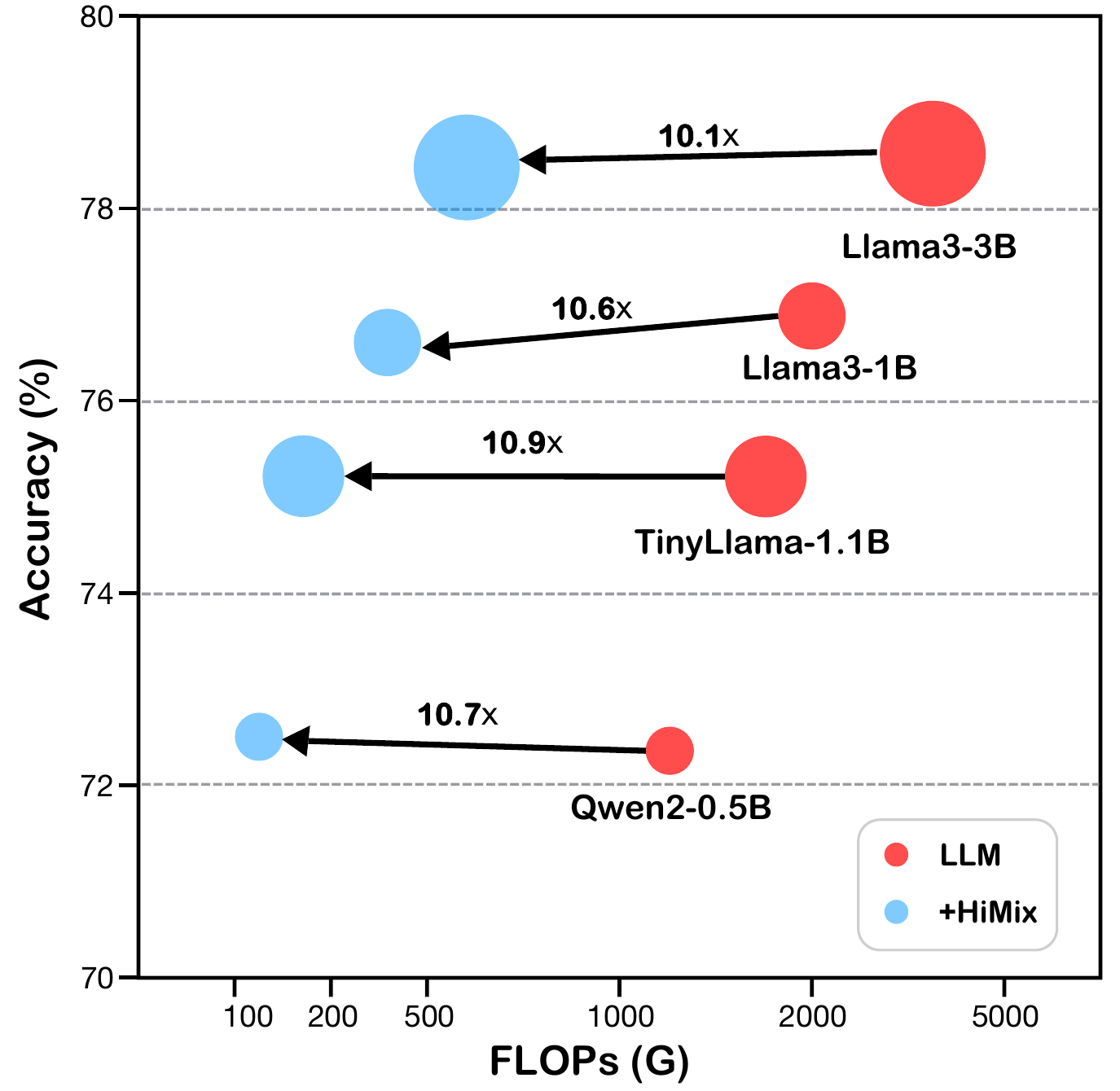}
    \caption{\textbf{Comparison of performance and computational cost of the language decoder between the original and HiMix models.} The circles, arranged from smallest to largest, represent the models Qwen2-0.5B, Llama3-1B, TinyLlama-1.1B, and Llama3-3B. While maintaining a similar performance to the original models, our HiMix achieves a \textbf{\bm{$10\times$}} reduction in computational cost.}
    \label{fig-flops}
\end{figure}

\begin{quote}
  \begin{flushleft}
    \textit{"What I hear, I forget. What I see, I remember."}
  \end{flushleft}
  \begin{flushright}
    \textbf{\hspace{4cm}---------} \textit{Confucius} % Adjust the length of the line as needed
  \end{flushright}
\end{quote}

% \begin{figure}[htb]
%     \centering
%     \includegraphics[width=\linewidth]{figure/fig-cosine_similarity.png}
%     \caption{Cosine similarity between hidden layer outputs and original input sequence across layers in the LLM and vision encoder.}
%     \label{fig-similarity}
% \end{figure}

% \begin{figure}[htb]
%     \centering
%     \includegraphics[width=\linewidth]{figure/fig-flops.png}
%     \caption{Comparison of Computational Cost and Performance: HiMix vs. Original Models}
%     \label{fig-similarity}
% \end{figure}

Understanding complex content is a pivotal step toward artificial general intelligence (AGI). As the two core modalities of information processing, vision and language each have unique advantages: \textcolor{ForestGreen}{\textit{text}} can provide detailed semantic information, while \textcolor{orange}{\textit{image}} can present intuitive vision cues. How to effectively integrate information from these two modalities has become a crucial orientation of current research. 
We have witnessed tremendous research and effort to achieve more complex and higher levels of semantic understanding built upon multimodal information integration~\cite{liu2024visual,li2022blip,li2023blip,zhu2023minigpt,Awadalla2023OpenFlamingoAO,Dai2023InstructBLIPTG,Gong2023MultiModalGPTAV}. Most models integrate vision and language features by concatenating them for input into Large Language Model (LLM)~\cite{achiam2023gpt,dubey2024llama,chiang2023vicuna}. However, vision sequences are typically longer than language sequences, significantly increasing computational complexity. This prompts us to wonder: \textit{Is simple concatenation truly the most efficient approach for facilitating interaction between vision and language?}

% Inspired by the famous quote of \textit{Confucius}, we begin to reassess the relative importance of vision and language. While image can aid humans in enhancing their memory, it is not necessary for models to rely on image  throughout the entire process. 
% % we argue that vision conveys information more efficiently than language.
Inspired by the famous quote of \textit{Confucius}, we begin to consider the role of visual and language information in the language decoder of LVLMs. We hypothesize that, during the forward propagation process, vision information may not need to be involved throughout the entire process.
A classic example is the story about the Treasure Hunt: the protagonist in the story typically has text information in the form of a \textcolor{ForestGreen}{\textit{riddle}} and image information in the form of a \textcolor{orange}{\textit{map}}. During the treasure hunt, the protagonist usually focuses on understanding the instructions in the riddle, consulting the map only when necessary to determine the direction. This analogy suggests that, as the backbone of the LVLM, the language model should primarily focus on \textcolor{ForestGreen}{\textit{text}} for comprehension and processing tasks, referring to \textcolor{orange}{\textit{image}} only when additional support is required.

\begin{figure}[t]
    \centering
    \includegraphics[width=\linewidth]{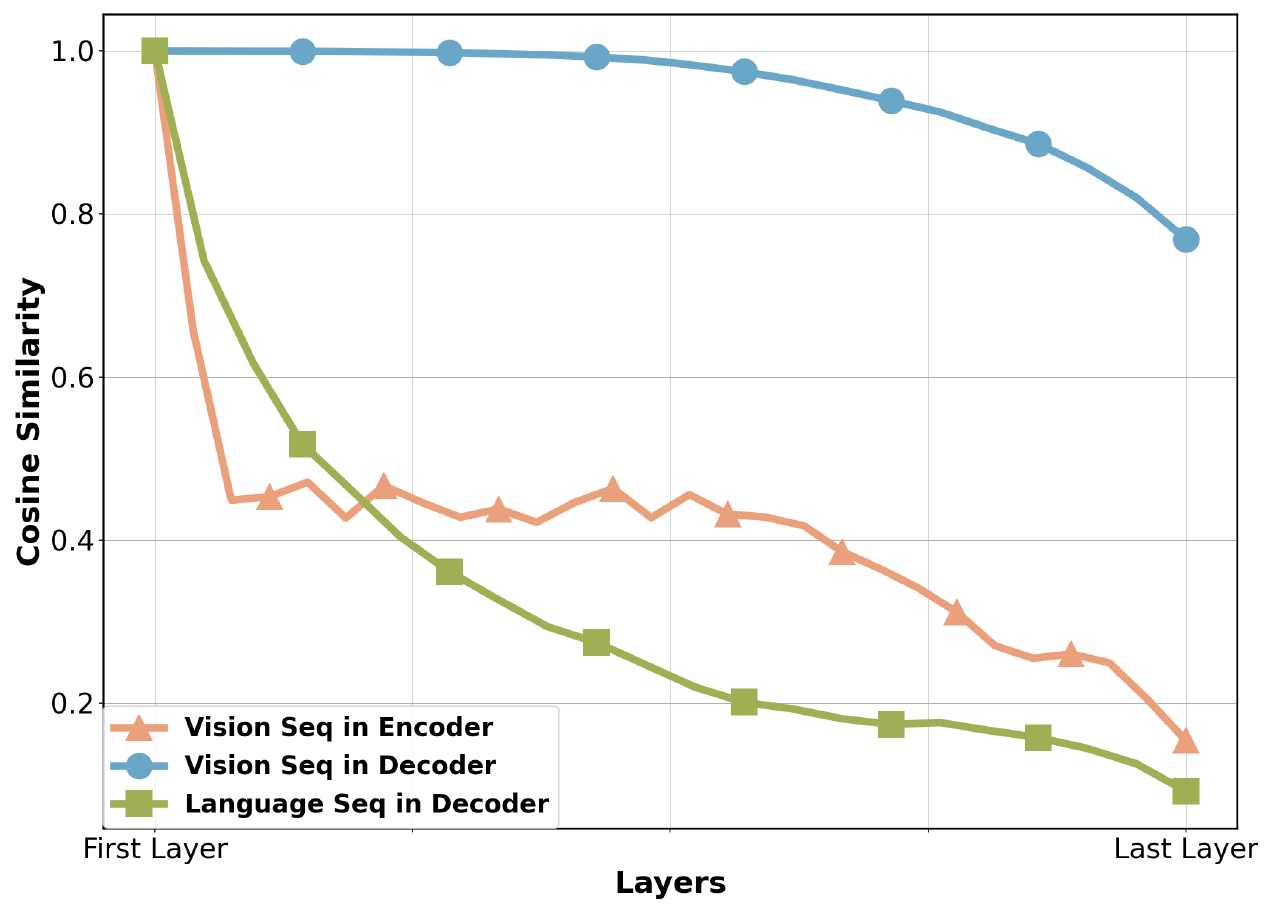}
    \caption{Layer-wise Cosine Similarity of Vision and Language Sequences.}
    \label{fig-sim}
\end{figure}

Therefore, we further analyze the layer-wise divergences generated during the forward propagation of vision and language sequences to support this hypothesis. Specifically, we examine the divergences between the original sequences input to the language decoder and the vision encoder of the LVLM, and the output sequences of each layer. The divergences are measured by cosine similarity, as shown in Figure~\ref{fig-sim}. It is obvious that in the vision encoder, the divergence of vision sequences shows significant changes from shallow to deep layers. This implies that as the model deepens, the vision encoder can capture higher-level semantic information. However, in the language decoder, the divergence between the vision sequences and the original sequences is much smaller compared to the language sequences. This reveals that vision sequences are updated less during forward propagation and are not the primary focus of the model, which undoubtedly supports our hypothesis.
We boldly speculate that it is feasible to remove vision sequences from the forward propagation in the language decoder of the LVLM. By incorporating these sequences only at specific stages, we can achieve comparable performance while significantly reducing overall computational complexity.

Based on these insights, we introduce \textbf{H}ierarchical Vision \textbf{i}njection for \textbf{Mix}ture Attention (\textbf{\textcolor{orange}{Hi}\textcolor{ForestGreen}{Mix}}), a novel vision-language interaction mechanism for MLLMs.
Our method avoids concatenating vision and language sequences as input. Instead, only the language sequence is involved in the entire forward propagation, while the vision sequence is injected at specific stages within each layer to interact with the language through Mixture Attention. This approach allows the model to receive more diverse vision inputs while significantly reducing overall computational complexity. As shown in Figure~\ref{fig-flops}, HiMix reduces the computational cost to approximately $10\times$ lower than that of the original model across various base models, while maintaining comparable performance, highlighting the significant advantage of our method in boosting computational efficiency. The main contributions of our work are as follows: 

\begin{itemize}

% \item We hypothesize that vision information does not need to be involved throughout the entire forward propagation process.

\item We hypothesize that visual information is not necessary throughout the forward propagation process in the language decoder of the LVLM.

% \item We propose a novel vision-language interaction mechanism where only the language sequence progresses through the full forward propagation, while vision sequences are injected at specific stages to enhance the model's performance.

\item We introduce a novel vision-language interaction mechanism in which only the language sequence undergoes full forward propagation, while vision sequences are injected at specific stages to enhance the model's performance.

% \item We demonstrate that HiMix maintains competitive performance while significantly reducing computational complexity, achieving robust performance on various models, such as Qwen2 and Llama3.

\item Our HiMix maintains competitive performance on various models, \textit{e.g.}, Qwen2 and Llama3, while achieving $10\times$ reductions on computational complexity.

\end{itemize}
\section{Related Works}
\label{sec:Related Works}
\subsection{Large Vision-Language Models}

With the remarkable performance of LLMs~\cite{achiam2023gpt,touvron2023llama,team2024gemma,zeng2022glm,bai2023qwen,zhang2024tinyllama} in natural language processing, researchers have extended these models to the multimodal domain, enabling more advanced vision and language understanding tasks. The typical architecture of LVLMs~\cite{liu2024visual,zhu2023minigpt,Bai2023QwenVLAF,chen2024internvl,zhou2024tinyllava} includes three main components: a visual encoder, a vision-language connector, and a large language model. Most approaches focus on aligning visual features with the textual space effectively. In this field, Flamingo~\cite{alayrac2022flamingo} pioneers the use of resamplers and cross-attention dense blocks to integrate visual features into LLMs. Following this, BLIP-2~\cite{li2023blip} introduces Qformer, which leverages frozen image encoders and LLMs to perform various vision-language tasks. The LLaVA series~\cite{liu2024visual,Li2024LLaVAOneVisionEV,Liu2023ImprovedBW} and MiniGPT-4~\cite{zhu2023minigpt} apply simple but effective projection layers to align different modalities and improved instruction-following abilities through multimodal instruction tuning data. Fuyu~\cite{fuyu-8b} aligns image patches to the LLM embeddings through a simple linear mapping. InternVL~\cite{chen2024internvl} introduces a large-scale vision encoder that integrates with QLLaMA as middleware for alignment with the LLM. Qwen-VL~\cite{Bai2023QwenVLAF} proposes a Position-aware Vision-Language Adapter, which compresses visual feature sequences to a fixed length before input to the LLM.

However, as model sizes continue to increase, the computational resources required also grow, calling for more efficient input processing methods to manage the increasing contextual demands.

\subsection{Vision Token Reduction}

The primary reason for the high computational cost of LVLMs is the excessive length of input sequences, often due to numerous prefix visual tokens. As computational costs scale quadratically with the number of input tokens, researchers have explored methods to reduce visual tokens and thereby lower computational expenses effectively.

FastV~\cite{chen2025image} is among the first to identify the inefficiencies associated with visual token usage in LVLMs. This model uses signals from the LLM to guide visual token pruning, trimming half of the image tokens within intermediate layers of the LLM. VTW~\cite{lin2024boosting} takes this further by completely removing all visual tokens after a certain intermediate layer, relying only on text tokens for subsequent computations. LLaVA-PruMerge~\cite{shang2024llava} dynamically retains visual tokens based on their similarity, effectively compressing the number of visual tokens. \( M^3 \)~\cite{cai2024matryoshka}, inspired by Matryoshka Representation Learning, learns multi-granular, coarse-to-fine token representations to control the number of tokens involved in computation.

While these approaches effectively reduce computational loads, they primarily focus on pruning or fully excluding visual tokens from certain computations. In contrast, our HiMix method avoids involving vision sequences in the forward propagation process. Rather than merely trimming or excluding visual tokens, HiMix hierarchically injects vision information at specific stages, allowing the model to reference visual cues only when necessary. This approach significantly reduces computational cost while preserving model performance.
\section{Method}
\label{sec:Method}

\begin{figure*}[h]
    \centering
    \includegraphics[width=\linewidth]{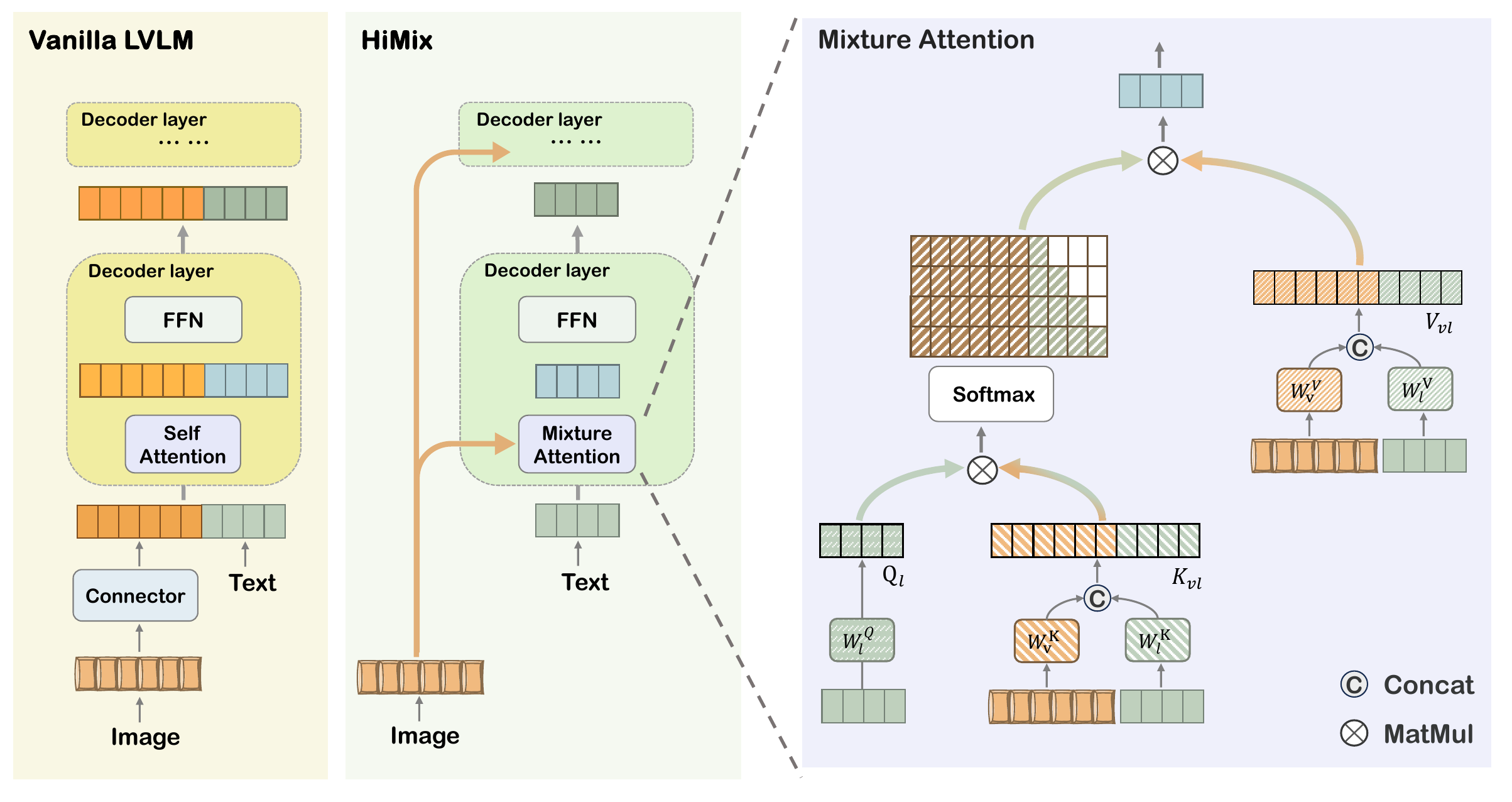}
    \caption{Comparison of Vanilla Model and HiMix Architectures. \textbf{Left:} Overall structure of traditional Vanilla. \textbf{Middle:} Overall structure of HiMix. \textbf{Right:} Details of HiMix.}
    \label{fig-Overall}
\end{figure*}

This paper presents a novel approach to vision-language understanding by modifying the interaction between multimodal information. In the preliminary section, we briefly introduce the commonly used LVLM method, referred to as the Vanilla-LVLM, followed by a detailed description of our proposed HiMix structure.

\subsection{Preliminary}
For an input image $I$, we extract the vision feature sequence $X_v = f_{\text{VE}}(I)$ using a pre-trained vision encoder. In the Vanilla-LVLM, these vision features are transformed through a connector to align with the language features, producing a sequence $X_{v'} \in \mathbb{R}^{N \times d}$, where $N$ is the length of the vision sequence and $d$ is the feature dimension.
For the input text $T$, features are extracted using a text encoder bonded to the LLM: $X_l = f_{\text{TE}}(T)$, where $X_l \in \mathbb{R}^{M \times d}$ denotes the language feature sequence, with $M$ as the length of the language sequence.

In the Vanilla-LVLM, the concatenation of the vision and language sequences is fed into the LLM: 
\begin{equation}
    X_{vl} = \left[X_{v'}; X_l\right]
\end{equation}

Each Transformer layer comprises two sub-layers: a self-attention layer and a feed-forward network (FFN). The attention layer includes three linear transformation matrices, $W^Q$, $W^K$, and $W^V$, which transform the input $X_{vl}$ into $Q$, $K$, and $V$ matrices. Subsequently, self-attention enables interaction within the sequence:
\begin{equation}
\text{SA} = \text{Softmax}\left(\frac{QK^T}{\sqrt{d}}\right)V
\end{equation}
A causal mask is used during attention to prevent information leakage from future tokens.
Following self-attention, the FFN typically consists of two linear transformation layers with a nonlinear activation function. The output $A$ of the self-attention layer is fed into the FFN as input:
\begin{equation}
 \text{FFN} = W_2(\text{Act}(W_1A))
\end{equation}
where $W_1$ and $W_2$ are the weight matrices for linear transformations. The overall architecture of the Vanilla-LVLM is shown on the left side of Figure~\ref{fig-Overall}.

For an input sequence of length $N + M$, the computational complexity of the Vanilla Model includes two parts: self-attention requires $O((N + M)^2 d)$ due to the $QK^T$ operation, while the feed-forward network, with two linear layers, contributes $O(8(N + M) d^2)$. Thus, the total complexity is:
\begin{equation}
    O((N+M)^2d)+O(8(N+M)d^2)
\end{equation}
Since the vision sequence length $N$ is usually longer than the language sequence length $M$, reducing the number of vision tokens in computation can effectively decrease overall computational costs.

\subsection{HiMix}
We introduce \textbf{H}ierarchical Vision \textbf{i}njection for \textbf{Mix}ture Attention (HiMix), designed to reduce computational complexity while maintaining LVLM performance. The key is that the vision sequence does not need to participate in the entire forward propagation process. In this section, we describe the design of Hierarchical Vision Injection and the implementation of Mixture Attention. The middle part of Figure~\ref{fig-Overall} illustrates the overall structure of HiMix, while the right side shows the detailed implementation of Mixture Attention.

\subsubsection{Hierarchical Vision Injection}
\begin{figure}[h]
    \centering
    \includegraphics[width=\linewidth]{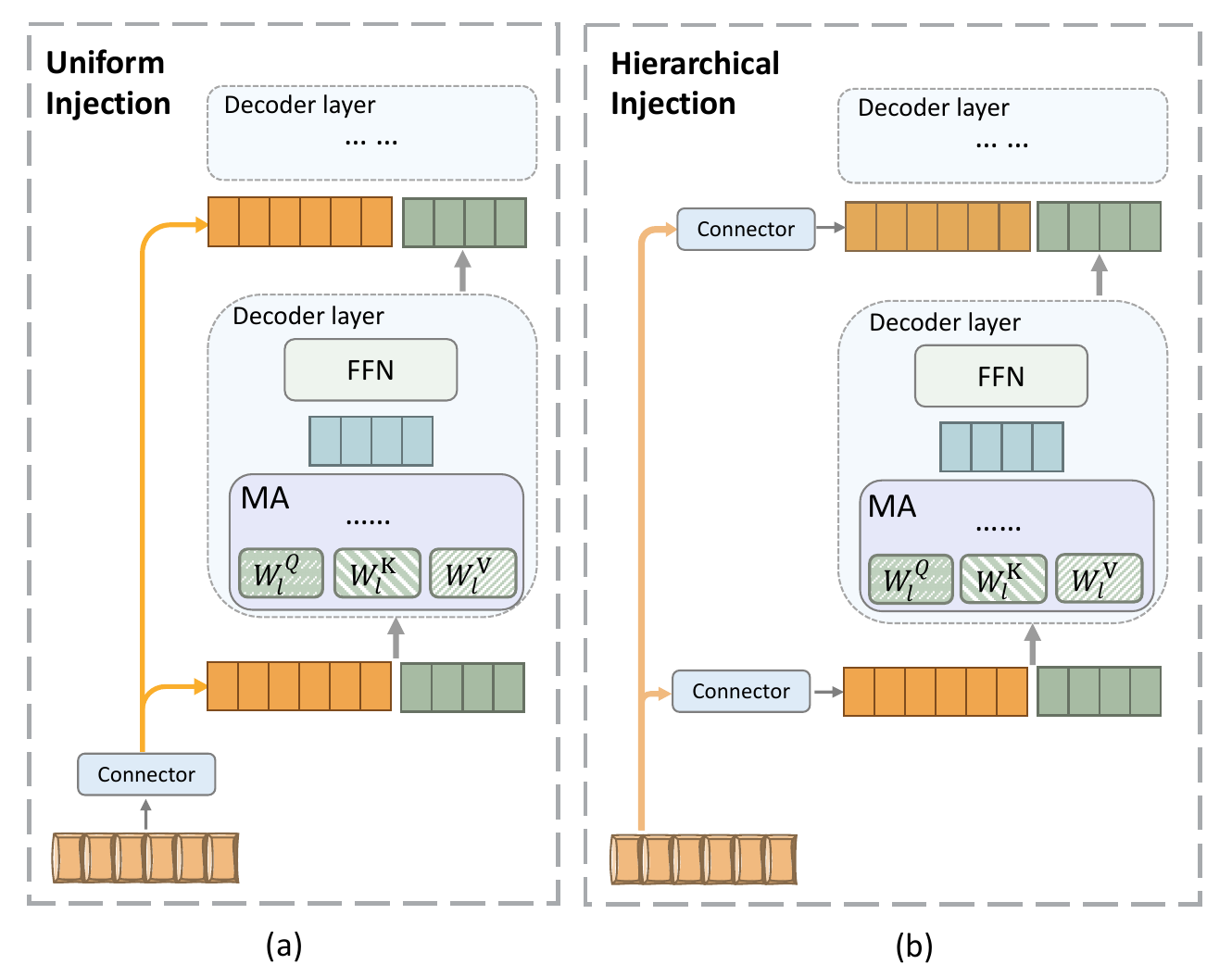}
    \caption{Exploration of Model Architecture Design. (a) Uniform Vision Injection to Each Layer. (b) Hierarchical Vision Injection Through Multi-Level Connectors. The Mixture Attention (MA) differs from the main method in that the vision and language sequences share the KV projection layers}
    \label{fig-evolution}
\end{figure}
We assume the vision sequence only needs to be injected when necessary, without being passed in layers.

A straightforward approach is to directly feed $X_{v'}$ into every layer, as shown in Figure~\ref{fig-evolution} (a), which represents uniform vision injection. However, this may constrain the model’s ability to effectively capture diverse vision information.

To provide the model with more diverse vision cues, an improved strategy is to increase the number of connectors to match the LLM layers, as illustrated in Figure~\ref{fig-evolution} (b). The vision sequence $X_v$ is first transformed by a corresponding connector before being fed into each layer, achieving hierarchical vision injection. 

However, these structures have a common issue: in subsequent attention computations, the vision and language sequences share the same KV projection layers. This may not be the most optimal solution. Since the language sequence undergoes attention and passes through the feed-forward network (FFN) at each layer, sharing the same KV projections with the vision sequence could introduce interference between the two modalities.

In our final model, we address this issue by merging each connector with the projection layers, resulting in dedicated vision projection layers. These layers not only align the dimensions between the modalities but also enable the hierarchical injection of vision features in the most straightforward manner. Finally, we send the vision sequence $X_v$ to each layer and integrate the vision projection layers into the attention mechanism, ensuring no interference between the vision and language sequences.
% However, this design results in a more complex model structure, which may not be the most efficient.

% In our final model, we merge each connector with the projection layers in the attention mechanism, creating vision projection layers. This integration serves dual purposes: dimensional alignment and hierarchical feature injection, while also preventing interference between vision and language features that could occur if they shared the same projection layer.

\subsubsection{Mixture Attention}
In Mixture Attention, the vision sequence $X_v$ and language sequence $X_l$ are treated as two separate inputs. 

Initially, the inputs are processed through five linear transformation matrices. Vision features are transformed through $W_{v}^{K}$ and $W_{v}^{V}$ to produce matrices $K_v$ and $V_v$. Language features are transformed via $W_{l}^{Q}$, $W_{l}^{K}$, $W_{l}^{V}$ to produce matrices $Q_l$, $K_l$, and $V_l$. 

We then concatenate the vision and language information to form the complete \textbf{KV} sequence:
\begin{equation}
K_{vl} = \left[K_v; K_l\right], \quad V_{vl} = \left[V_v; V_l\right]
\end{equation}
Subsequently, mixture attention facilitates multimodal information interaction:
\begin{equation}
\text{MA}= \text{Softmax}\left(\frac{Q_l{K_{vl}}^T}{\sqrt{d}}\right)V_{vl}
\end{equation}

The resulting sequence maintains the same length as the initial language feature sequence, and the vision sequence does not require forward propagation.

During the attention calculation, positional embeddings are added to $Q_l$ and $K_l$. Note that since the vision features already contain positional information from feature extraction, we do not add additional positional embedding to the vision sequence. 

We apply a part-causal mask to the attention weights: all language tokens can attend to previous vision tokens, while a causal mask is applied for language interactions to prevent future information leakage.

For an input sequence of length $N+M$, the computational complexity of HiMix includes two parts: mixture attention requires $O((N+M)Md)$ due to the $Q_l{K_{vl}}^T$ operation,  while the feed-forward network, applied only to the language sequence, contributes $O(8Md^2)$. Thus, the total complexity is:
\begin{equation}
    O((N+M)Md)+O(8Md^2)
\end{equation}
Compared to the complexity of the Vanilla-LVLM, HiMix significantly reduces computational cost. By avoiding the quadratic term associated with self-attention across both sequences, HiMix provides a more efficient solution, especially advantageous for handling longer vision sequences.

The complete algorithmic workflow of HiMix is presented in Algorithm ~\ref{alg:hdsa}.

% \begin{algorithm}
% \caption{HiMix: Hierarchical Vision Injection for Mixture Attention}
% \label{alg:hdsa}
% \begin{algorithmic}[1]
% \State \textit{Input:} Image $I$, Text $T$
% \State \textit{Output:} Language sequence $Y_l$ after $l$ layers of processing
% \State \textit{Initialize:} 
% \State \quad $X_v \gets f_{\text{VE}}(I)$
% \State \quad $X_l \gets f_{\text{TE}}(T)$
% \State \quad $Y_0 \gets X_l$
% \For{$i = 1$ to $l$}
%     \State \textit{Language Projection:}
%     \State \quad $Q_l^i, K_l^i, V_l^i \gets \textit{Proj}_l(Y_{i-1})$
%     \State \textit{Vision Projection:}
%     \State \quad $K_v^i, V_v^i \gets \textit{Proj}_v(X_v)$
%     \State \textit{Add Positional Embedding:}
%     \State \quad $Q_{l'}^{i} \gets Q_l^i + P_l$, $K_{l'}^{i} \gets K_l^i + P_l$
%     \State \textit{Concatenate Sequences:}
%     \State \quad $K_{vl}^i \gets [K_v^i; K_{l'}^{i}]$, $V_{vl}^i \gets [V_v^i; V_l^i]$
%     \State \textit{Calculate Attention Scores:}
%     \State \quad $\text{S}^i \gets (Q_{l'}^{i} (K_{vl}^i)^T) \text{/}\sqrt{d}$
%     \State \textit{Apply Part-Causal Mask:}
%     \State \quad $\hat{\text{S}}^i \gets \text{S}^i + \text{Mask}$
%     \State \textit{Compute Attention Output:}
%     \State \quad $\text{A}_i \gets \textit{Softmax}(\hat{\text{S}}^i) V_{vl}^i$
%     \State \textit{Feed-forward Network:}
%     \State \quad $\text{Y}_i \gets W_2(\text{Act}(W_1 \text{A}_i))$  
% \EndFor
% \State \Return $Y_l$
% \end{algorithmic}
% \end{algorithm}

\begin{algorithm}
\caption{HiMix: Hierarchical Vision Injection for Mixture Attention}
\label{alg:hdsa}
\begin{algorithmic}[1]
\State \textbf{Input:} Image $I$, Text $T$
\State \textbf{Output:} Language sequence $Y_l$ after $l$ layers of processing
\State \textbf{Initialize:} 
\State \quad $X_v \gets f_{\text{VE}}(I)$
\State \quad $X_l \gets f_{\text{TE}}(T)$
\State \quad $Y_0 \gets X_l$
\For{$i = 1$ to $l$}
    \State \textbf{ Language Projection: }
    \State \quad $Q_l^i, K_l^i, V_l^i \gets \textit{Proj}_l(Y_{i-1})$
    \State \textbf{Vision Projection:}
    \State \quad $K_v^i, V_v^i \gets \textit{Proj}_v(X_v)$
    \State \textbf{  Add Positional Embedding: }
    \State \quad $Q_{l'}^{i} \gets Q_l^i + P_l$, $K_{l'}^{i} \gets K_l^i + P_l$
    \State \textbf{  Concatenate Sequences: }
    \State \quad $K_{vl}^i \gets [K_v^i; K_{l'}^{i}]$, $V_{vl}^i \gets [V_v^i; V_l^i]$
    \State \textbf{  Calculate Attention Scores: }
    \State \quad $S^i \gets (Q_{l'}^{i} (K_{vl}^i)^T) \text{/}\sqrt{d}$
    % \State \quad $\text{S}^k \gets \frac{Q_{l'}^{k} (K_{vl}^k)^T}{\sqrt{d}}$
    \State \textbf{  Apply Part-Causal Mask: }
    \State \quad $\hat{S}^i \gets S^i + Mask$
    \State \textbf{  Compute Attention Output: }
    \State \quad $A_i \gets \textit{Softmax}(\hat{S}^i) V_{vl}^i$
    \State \textbf{ Feed-forward Network: }
    \State \quad $Y_i \gets W_2(\textit{Act}(W_1 A_i))$  
\EndFor
\State \Return $Y_l$
\end{algorithmic}
\end{algorithm}

\section{Experiments}

In this section, we first introduce the experimental setup, followed by a detailed discussion of the ablation study, analyzing the impact of each component on model performance and efficiency. Next, we present the main experimental results, demonstrating the effectiveness of the proposed HiMix method across multiple benchmarks. Finally, we conduct additional extended experiments.

\subsection{Experimental Setup}

\textbf{Model Selection.}
We select TinyLlama~\cite{zhang2024tinyllama}, Qwen2~\cite{yang2024qwen2}, and Llama3.2~\cite{dubey2024llama} as language decoders, with the pre-trained Siglip~\cite{zhai2023sigmoid} vision encoder SoViT-400m/14 as the vision encoder. In the vanilla-LVLM setup, a 2-layer MLP serves as the connector between the vision encoder and language decoder, similar to the configuration in TinyLLaVA~\cite{zhou2024tinyllava}. During initialization, the corresponding pre-trained weights are loaded, and the parameters of the vision projection layers are randomly initialized. 

\noindent\textbf{Evaluation Metrics.}
To evaluate our proposed HiMix method, we assess both model efficiency and performance metrics. The experiments are designed to validate whether HiMix could significantly reduce computational complexity while maintaining competitive performance.

For model efficiency, we report the number of model parameters and computational costs (in GFLOPs). Each input sample contains an image with 728 vision tokens processed by the Siglip encoder and a text prompt sequence of length 64. This setup simulates a typical multimodal input scenario found in real-world applications.

For model performance, we evaluate the model on seven commonly used benchmarks: VQAv2~\cite{goyal2017vqav2}, GQA~\cite{hudson2019gqa}, TextVQA~\cite{singh2019textvqa}, MM-Vet~\cite{Yu2023MMVetEL}, POPE~\cite{li2023pope}, MME~\cite{Fu2023MMEAC}, and MMMU~\cite{yue2024mmmu}. These benchmarks comprehensively measure model capabilities, ranging from basic vision perception to advanced reasoning.

\noindent\textbf{Training Strategies.}
We adopt two distinct training strategies, referred to as "Regular Paradigm" and "Enhanced Paradigm":

\begin{itemize} 
    \item[-] \textit{Regular Paradigm.}This strategy uses the LLaVA-1.5~\cite{liu2024visual} dataset for training. The model is first pretrained with LLaVA-1.5-558k, where the vision encoder and language decoder parameters are kept frozen, and only the vision projection layers are trained. In the supervised fine-tuning (SFT) stage, the model is fine-tuned with LLaVA-1.5-mix-665k, keeping the vision encoder frozen but updating all parameters of the language decoder.
    
    \item[-] \textit{Enhanced Paradigm.}This strategy leverages both the LLaVA-1.5~\cite{liu2024visual} and ShareGPT4V~\cite{chen2023sharegpt4v} datasets to further enhance performance, following a three-stage training approach inspired by TinyLLaVA~\cite{zhou2024tinyllava}. The initial pretraining stage mirrors the setup in the Regular Paradigm. Next, the model undergoes additional pretraining with the ShareGPT4V-pretrain-1246k dataset, followed by fine-tuning with the ShareGPT4V-mix-665k dataset in the SFT stage. Both the additional pretraining and SFT stages use the same configuration as the SFT stage in the Regular Paradigm. \end{itemize}

For the main experiments, both training strategies are applied, while the ablation studies use the Regular Paradigm."

\subsection{Ablation Study}

\textbf{Self-Attention vs Mixture Attention.}
We first examine the impact on model efficiency when Self-Attention in the language decoder is replaced with Mixture Attention, the overall structure illustrated in Figure~\ref{fig-evolution}(a). Siglip is used as the vision encoder, and Llama3-1B is used as the language decoder. As shown in Table~\ref{tab-abla_attn}, without increasing parameter count, Mixture Attention significantly reduces computational overhead. The overall computational cost (GFLOPs) of HiMix is only 33\% of that of the original model, while the computational cost of the language decoder alone is reduced to  approximately 10\% of the original. These results demonstrate that Mixture Attention enables efficient multimodal information interaction, greatly reducing computational costs.
\setlength{\tabcolsep}{5pt}
\begin{table}[htp]
\caption{Efficiency Comparison between Models with Self-Attention and Mixture Attention. S represents Siglip. L represents Llama3-1B.}
\label{tab-abla_attn}
\centering
\resizebox{\linewidth}{!}{
\begin{tabular}{ccccccc}
\hline
                         & \multicolumn{2}{c}{Overall  }                                                                                                    & \multicolumn{2}{c}{Language Decoder}                                                                                           & \multicolumn{2}{c}{Vision Encoder}                                                                                             \\
\multirow{-2}{*}{Method} & \multicolumn{1}{c}{\begin{tabular}[c]{@{}c@{}}Params\\ (B)\end{tabular}} & \begin{tabular}[c]{@{}c@{}}FLOPs\\ (G)\end{tabular} & \multicolumn{1}{c}{\begin{tabular}[c]{@{}c@{}}Params\\ (B)\end{tabular}} & \begin{tabular}[c]{@{}c@{}}FLOPs\\ (G)\end{tabular} & \multicolumn{1}{c}{\begin{tabular}[c]{@{}c@{}}Params\\ (B)\end{tabular}} & \begin{tabular}[c]{@{}c@{}}FLOPs\\ (G)\end{tabular} \\ \hline
S+L               & \multicolumn{1}{c}{1.67}                                                 & 2720                                                & \multicolumn{1}{c}{1.24}                                                 & 2040                                                & \multicolumn{1}{c}{0.43}                                                 & 671                                                 \\ 
\textit{w.}MA                     & \multicolumn{1}{c}{1.67}                                                 & {\color[HTML]{1F2328} \textbf{894(33\%)}}           & \multicolumn{1}{c}{1.24}                                                 & {\color[HTML]{1F2328} \textbf{214(10\%)}}           & \multicolumn{1}{c}{0.43}                                                 & 671                                                 \\ \hline
\end{tabular}
}
\end{table}

% \noindent\textbf{Fixed vs. Hierarchical vision Features} 
\noindent\textbf{Uniform Injection vs Hierarchical Injection.}
We next investigate the impact of different vision injection strategies on model performance by comparing uniform vision injection with hierarchical injection. As presented in Table~\ref{tab-abla_feature}, uniform injection of vision information across all layers (illustrated in Figure~\ref{fig-evolution}(a)) leads to a decline in model performance, while hierarchical injection (illustrated in Figure~\ref{fig-evolution}(b)) improves overall performance. Although hierarchical injection incurs slightly higher computational cost compared to uniform injection, it still reduces the overall computational cost by nearly 10x compared to the original approach. Given that hierarchical injection provides more diverse vision information, it enables the model to better handle more complex tasks. This demonstrates that a hierarchical structure is a more effective approach, striking a better balance between performance and computational efficiency.

\setlength{\tabcolsep}{5pt}
\begin{table}[htp]
\caption{Performance Comparison Between Uniform and Hierarchical Vision Injection.}
\label{tab-abla_feature}
\resizebox{\linewidth}{!}{
\renewcommand{\arraystretch}{1.3} % Increase the row height (default is 1)
\begin{tabular}{ccccccc}
\hline
Decoder                      & VQA\textsuperscript{v2} & GQA  & VQA\textsuperscript{T} &  POPE & MME    & FLOPs(G) \\ \hline
Llama3-1B                    & 76.8   & 59.6 & 52.7     & 86.7 & 1335.0   & 2030              \\ 
\textit{w.}Uniform     & 74.9   & 57.8 & 49.7   & 85.8 & 1243.9 & 201               \\ 
\textit{w.}Hierarchical & 77.3   & 60.7 & 50.4    & 85.9 & 1260.4 & 354               \\ \hline
\end{tabular}
}
\end{table}

\subsection{Comprehensive Assessment}
\begin{table*}[!htp]
\caption{Comprehensive comparison of performance and computational efficiency between the original model and HiMix \textbf{under the Regular Paradigm}, using SigLIP as the vision encoder. Performance metrics include VQAv2, GQA, TextVQA, MM-Vet, POPE, MME, and MMMU. Computational efficiency is assessed by Language Decoder parameter count (B) and GFLOPs. Performance improvements are highlighted in \textbf{bold}, with computational cost shown in \textbf{\textcolor[HTML]{FF0000}{red}} as a percentage of the original model.}
\label{tab-main_2Stage}
\makebox[\textwidth][c]{ 
\resizebox{\textwidth}{!}{
\begin{tabular}{cccccccccc}
\hline
                           & \multicolumn{7}{c}{Performances}                                                                                                                                                                                                                                                                                                                                                                                                                                                                           & \multicolumn{2}{c}{Efficiency}                                                                                                 \\
\multirow{-2}{*}{\begin{tabular}[c]{@{}c@{}}Language\\ Decoder\end{tabular}}& \multicolumn{1}{c}{VQA\textsuperscript{v2}~\cite{goyal2017vqav2}}                               & \multicolumn{1}{c}{GQA~\cite{hudson2019gqa}}                                  & \multicolumn{1}{c}{VQA\textsuperscript{T}~\cite{singh2019textvqa}}                     & \multicolumn{1}{c}{MM-Vet~\cite{Yu2023MMVetEL}}                               & \multicolumn{1}{c}{POPE~\cite{li2023pope}}                        & \multicolumn{1}{c}{MME~\cite{Fu2023MMEAC}}                           & MMMU~\cite{yue2024mmmu}                                & \multicolumn{1}{c}{\begin{tabular}[c]{@{}c@{}}Params\\ (B)\end{tabular}} & \begin{tabular}[c]{@{}c@{}}FLOPs\\ (G)\end{tabular} \\ \hline
Qwen2-0.5B                                         & \multicolumn{1}{c}{72.3}                                 & \multicolumn{1}{c}{55.8}                                 & \multicolumn{1}{c}{45.2}                        & \multicolumn{1}{c}{19.5}                                 & \multicolumn{1}{c}{86.6}                        & \multicolumn{1}{c}{1153.0}                        & 29.7                                 & \multicolumn{1}{c}{{\color[HTML]{333333} 0.49}}                          & 837                                                 \\ 
+HiMix                                             & \multicolumn{1}{c}{{\textbf{72.6}}} & \multicolumn{1}{c}{{\textbf{55.9}}} & \multicolumn{1}{c}{44.2}                        & \multicolumn{1}{c}{{\textbf{21.1}}} & \multicolumn{1}{c}{84.9}                        & \multicolumn{1}{c}{1039.7}                        & {\textbf{31.6}} & \multicolumn{1}{c}{0.50}                                                 & { \textbf{78\color[HTML]{FE0000}(9\%)}}             \\ 
TinyLlama-1.1B                                     & \multicolumn{1}{c}{{\color[HTML]{1F2328} 75.5}}          & \multicolumn{1}{c}{{\color[HTML]{1F2328} 58.6}}          & \multicolumn{1}{c}{{\color[HTML]{1F2328} 49.6}} & \multicolumn{1}{c}{{\color[HTML]{1F2328} 23.5}}          & \multicolumn{1}{c}{{\color[HTML]{1F2328} 86.3}} & \multicolumn{1}{c}{{\color[HTML]{1F2328} 1256.5}} & {\color[HTML]{1F2328} 28.3}          & \multicolumn{1}{c}{1.10}                                                 & 1750                                                \\ 
+HiMix                                             & \multicolumn{1}{c}{{\textbf{75.5}}} & \multicolumn{1}{c}{{\textbf{59.2}}} & \multicolumn{1}{c}{44.1}                        & \multicolumn{1}{c}{{\textbf{24.5}}} & \multicolumn{1}{c}{85.7}                        & \multicolumn{1}{c}{1179.0}                        & {\textbf{29.0}} & \multicolumn{1}{c}{1.11}                                                 & { \textbf{161\color[HTML]{FE0000}(9\%)}}            \\
Llama-3.2-1B                                       & \multicolumn{1}{c}{76.8}                                 & \multicolumn{1}{c}{59.6}                                 & \multicolumn{1}{c}{52.7}                        & \multicolumn{1}{c}{27.8}                                 & \multicolumn{1}{c}{86.7}                        & \multicolumn{1}{c}{1334.5}                        & 30.6                                 & \multicolumn{1}{c}{1.24}                                                 & 2040                                                \\ 
+HiMix                                             & \multicolumn{1}{c}{76.2}                                 & \multicolumn{1}{c}{{\textbf{59.6}}} & \multicolumn{1}{c}{50.5}                        & \multicolumn{1}{c}{{\textbf{28.4}}} & \multicolumn{1}{c}{86.6}                        & \multicolumn{1}{c}{1198.6}                        & 29.9                                 & \multicolumn{1}{c}{1.25}                                                 & {\textbf{192\color[HTML]{FE0000} (9\%)}}            \\ 
Llama-3.2-3B                                       & \multicolumn{1}{c}{78.6}                                 & \multicolumn{1}{c}{62.5}                                 & \multicolumn{1}{c}{55.3}                        & \multicolumn{1}{c}{31.5}                                 & \multicolumn{1}{c}{86.9}                        & \multicolumn{1}{c}{1449.3}                        & 34.8                                 & \multicolumn{1}{c}{3.21}                                                 & 5310                                                \\ 
+HiMix                                             & \multicolumn{1}{c}{78.4}                                 & \multicolumn{1}{c}{61.4}                                 & \multicolumn{1}{c}{54.4}                        & \multicolumn{1}{c}{30.2}                                 & \multicolumn{1}{c}{86.6}                        & \multicolumn{1}{c}{1261.9}                        & {\textbf{35.6}} & \multicolumn{1}{c}{3.28}                                                 & {\textbf{525\color[HTML]{FE0000} (9\%)}}            \\ \hline

\end{tabular}
}
}
\end{table*}
\begin{table*}[!htp]
\caption{Comprehensive comparison of performance and computational efficiency between the original model and HiMix \textbf{under the Enhanced Paradigm}, using SigLIP as the vision encoder. Performance improvements are highlighted in \textbf{bold}, with computational cost shown in \textbf{\textcolor[HTML]{FF0000}{red}} as a percentage of the original model.}
\label{tab-main_3Stage}
\makebox[\textwidth][c]{ 
\resizebox{\textwidth}{!}{

\begin{tabular}{cccccccccc}
\hline
                                                                             & \multicolumn{7}{c}{Performances}                                                                                                                                                                                                                                                                                                               & \multicolumn{2}{c}{Efficiency}                                                                                                 \\ 
\multirow{-2}{*}{\begin{tabular}[c]{@{}c@{}}Language\\ Decoder\end{tabular}} & \multicolumn{1}{c}{VQA\textsuperscript{v2}~\cite{goyal2017vqav2}}                               & \multicolumn{1}{c}{GQA~\cite{hudson2019gqa}}                                  & \multicolumn{1}{c}{VQA\textsuperscript{T}~\cite{singh2019textvqa}}                     & \multicolumn{1}{c}{MM-Vet~\cite{Yu2023MMVetEL}}                               & \multicolumn{1}{c}{POPE~\cite{li2023pope}}                        & \multicolumn{1}{c}{MME~\cite{Fu2023MMEAC}}                           & MMMU~\cite{yue2024mmmu}                       & \multicolumn{1}{c}{\begin{tabular}[c]{@{}c@{}}Params\\ (B)\end{tabular}} & \begin{tabular}[c]{@{}c@{}}FLOPs\\ (G)\end{tabular} \\ \hline
Qwen2-0.5B                                                                   & \multicolumn{1}{c}{75.6}                        & \multicolumn{1}{c}{58.7}                        & \multicolumn{1}{c}{49.0}                        & \multicolumn{1}{c}{23.1}                        & \multicolumn{1}{c}{86.9}                        & \multicolumn{1}{c}{1255.6}                        & 31.2                        & \multicolumn{1}{c}{0.49}                                                 & 837                                                 \\
+HiMix                                                                       & \multicolumn{1}{c}{{\textbf{78.2}}} & \multicolumn{1}{c}{{\textbf{61.5}}} & \multicolumn{1}{c}{46.1}                        & \multicolumn{1}{c}{{\textbf{26.7}}} & \multicolumn{1}{c}{{\textbf{87.5}}} & \multicolumn{1}{c}{1162.1}                        & 29.7                        & \multicolumn{1}{c}{0.50}                                                 & { \textbf{78\color[HTML]{FE0000}(9\%)}}             \\ 
TinyLlama-1.1B                                                               & \multicolumn{1}{c}{{\color[HTML]{1F2328} 76.7}} & \multicolumn{1}{c}{{\color[HTML]{1F2328} 59.8}} & \multicolumn{1}{c}{{\color[HTML]{1F2328} 51.8}} & \multicolumn{1}{c}{{\color[HTML]{1F2328} 24.7}} & \multicolumn{1}{c}{{\color[HTML]{1F2328} 85.8}} & \multicolumn{1}{c}{{\color[HTML]{1F2328} 1271.6}} & {\color[HTML]{1F2328} 29.1} & \multicolumn{1}{c}{1.10}                                                 & 1750                                                \\ 
+HiMix                                                                       & \multicolumn{1}{c}{{\textbf{78.7}}} & \multicolumn{1}{c}{{\textbf{61.9}}} & \multicolumn{1}{c}{50.0}                        & \multicolumn{1}{c}{{\textbf{26.4}}} & \multicolumn{1}{c}{{\textbf{86.5}}} & \multicolumn{1}{c}{1261.7}                        & {\textbf{29.3}} & \multicolumn{1}{c}{1.11}                                                 & { \textbf{161 \color[HTML]{FE0000}(9\%)}}            \\
Llama-3.2-1B                                                                 & \multicolumn{1}{c}{78.2}                        & \multicolumn{1}{c}{60.9}                        & \multicolumn{1}{c}{53.8}                        & \multicolumn{1}{c}{29.2}                        & \multicolumn{1}{c}{86.6}                        & \multicolumn{1}{c}{1341.5}                        & 30.1                        & \multicolumn{1}{c}{1.24}                                                 & 2040                                                \\ 
+HiMix                                                                       & \multicolumn{1}{c}{{\textbf{79.1}}} & \multicolumn{1}{c}{{\textbf{61.3}}} & \multicolumn{1}{c}{51.9}                        & \multicolumn{1}{c}{{\textbf{29.9}}} & \multicolumn{1}{c}{{\textbf{86.8}}} & \multicolumn{1}{c}{1271.7}                        & 29.9                        & \multicolumn{1}{c}{1.25}                                                 & {\textbf{192\color[HTML]{FE0000} (9\%)}}            \\ 
Llama-3.2-3B                                                                 & \multicolumn{1}{c}{80.2}                        & \multicolumn{1}{c}{63.4}                        & \multicolumn{1}{c}{59.5}                        & \multicolumn{1}{c}{35.4}                        & \multicolumn{1}{c}{86.6}                        & \multicolumn{1}{c}{1452.6}                        & 36                          & \multicolumn{1}{c}{3.21}                                                 & 5310                                                \\ 
+HiMix                                                                       & \multicolumn{1}{c}{{\textbf{80.2}}} & \multicolumn{1}{c}{61.7}                        & \multicolumn{1}{c}{55.2}                        & \multicolumn{1}{c}{31.7}                        & \multicolumn{1}{c}{{\textbf{86.8}}} & \multicolumn{1}{c}{1343.8}                        & 31.7                        & \multicolumn{1}{c}{3.28}                                                 & { \textbf{525\color[HTML]{FE0000}(9\%)}}            \\ \hline
\end{tabular}
}
}
\end{table*}
\begin{table*}[!htp]
\caption{Efficiency Comparison of Baseline and HiMix across Different Input Scenarios. V:L represents the ratio of vision to language token lengths in the input sequence. FLOPs indicates the computational cost of the language decoder, and Max VRAM refers to the maximum GPU memory required during computation.}
\label{tab-effi}
\resizebox{\linewidth}{!}{
\begin{tabular}{cccccccccccc}
\hline
                         & \multicolumn{2}{c}{V:L 728:32}                                        & \multicolumn{2}{c}{V:L 728:64}                                        & \multicolumn{2}{c}{V:L 728:200}                                       & \multicolumn{2}{c}{V:L 728:728}                                       & \multicolumn{2}{c}{V:L 728:1000}                                      \\
\multirow{-2}{*}{Method} & \begin{tabular}[c]{@{}c@{}}FLOPs\\ (G)\end{tabular} & \begin{tabular}[c]{@{}c@{}}Max\\ VRAM(GB)\end{tabular} & \begin{tabular}[c]{@{}c@{}}FLOPs\\ (G)\end{tabular} & \begin{tabular}[c]{@{}c@{}}Max\\ VRAM(GB)\end{tabular} & \begin{tabular}[c]{@{}c@{}}FLOPs\\ (G)\end{tabular} & \begin{tabular}[c]{@{}c@{}}Max\\ VRAM(GB)\end{tabular} & \begin{tabular}[c]{@{}c@{}}FLOPs\\ (G)\end{tabular} & \begin{tabular}[c]{@{}c@{}}Max\\ VRAM(GB)\end{tabular} & \begin{tabular}[c]{@{}c@{}}FLOPs\\ (G)\end{tabular} & \begin{tabular}[c]{@{}c@{}}Max\\ VRAM(GB)\end{tabular} \\ \hline
Qwen2-0.5B               & 801                                                 & 7.41                                                   & 837                                                 & 7.56                                                   & 991                                                 & 8.32                                                   & 1620                                                & 11.95                                                  & 1970                                                & 14.22                                                  \\
\rowcolor[HTML]{D3D3D3} 
+HiMix                   & 44                                                  & 4.82                                                   & 78                                                  & 4.95                                                   & 224                                                 & 5.52                                                   & 821                                                 & 8.47                                                   & 1150                                                & 10.34                                                  \\
TinyLlama-1.1B           & 1680                                                & 9.15                                                   & 1750                                                & 9.35                                                   & 2080                                                & 10.65                                                  & 3400                                                & 16.95                                                  & 4120                                                & 21.04                                                  \\
\rowcolor[HTML]{D3D3D3} 
+HiMix                   & 92                                                  & 5.29                                                   & 161                                                 & 5.46                                                   & 466                                                 & 6.35                                                   & 1720                                                & 11.12                                                  & 2400                                                & 14.42                                                  \\
Llama-3.2-1B             & 1950                                                & 8.84                                                   & 2040                                                & 9.05                                                   & 2410                                                & 10.1                                                   & 3880                                                & 15.17                                                  & 4660                                                & 18.35                                                  \\
\rowcolor[HTML]{D3D3D3} 
+HiMix                   & 110                                                 & 5.44                                                   & 192                                                 & 5.6                                                    & 546                                                 & 6.38                                                   & 1970                                                & 10.32                                                  & 2730                                                & 12.96                                                  \\
Llama-3.2-3B             & 5080                                                & 14.89                                                  & 5310                                                & 15.24                                                  & 6260                                                & 16.83                                                  & 10090                                               & 24.28                                                  & 12130                                               & 28.92                                                  \\
\rowcolor[HTML]{D3D3D3} 
+HiMix                   & 310                                                 & 9.58                                                   & 525                                                 & 9.84                                                   & 1450                                                & 11.1                                                   & 5140                                                & 17.02                                                  & 7120                                                & 20.9                               \\ \hline                  
\end{tabular}
}
\end{table*}

\begin{figure*}[htp]
    \centering
    \includegraphics[width=\linewidth]{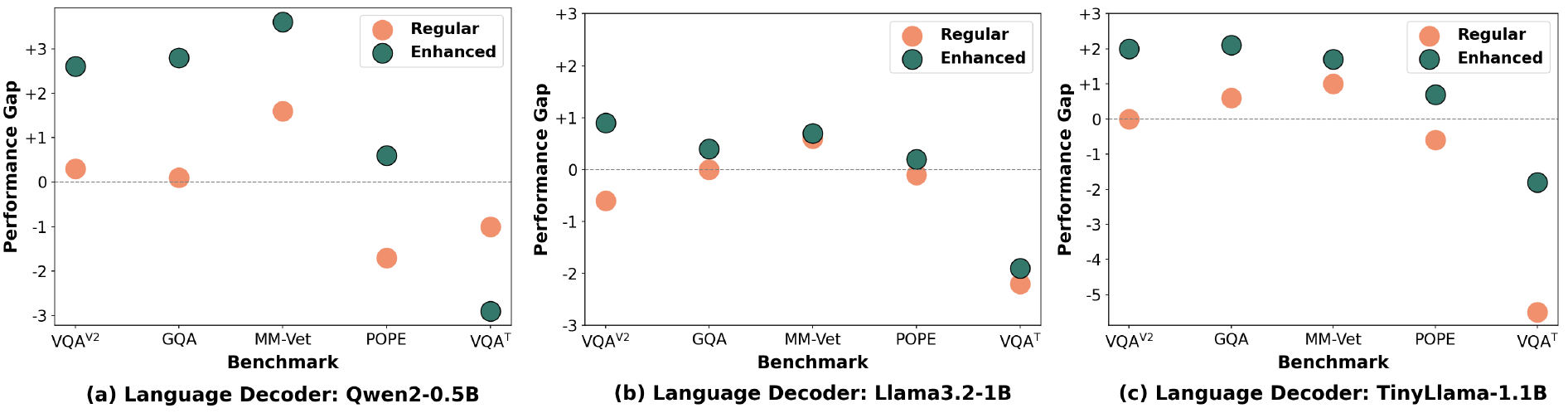}
    \caption{Performance gaps between HiMix and the baseline model across various benchmarks under two training strategies: \textcolor[HTML]{F0906D}{Regular Paradigm} and \textcolor[HTML]{46756B}{Enhanced Paradigm}. The y-axis represents the performance gap relative to the baseline, with positive values indicating improvements. Subfigures (a), (b), and (c) show results for Qwen2-0.5B, Llama3-1B, and TinyLLaMA-1.1B, respectively. Results indicate that using the Enhanced Paradigm amplifies HiMix's performance benefits over the baseline.}
    \label{fig-gap}
\end{figure*}

\begin{table*}[htp]
\caption{Performance and Parameters of Various Vision Encoders. All experiments use TinyLlama-1.1B (with HiMix applied) as the language decoder. InRes: Input Image Resolution; PatchSz: Patch Size; FeatDim: Vision Feature Dimensions (Token Count, Feature Dimension).}
\label{tab-abla_ve}
\resizebox{\linewidth}{!}{
\begin{tabular}{ccccccccccc}
\hline
\multirow{2}{*}{Vision Encoder} & \multicolumn{5}{c}{Model Parameters} & \multicolumn{5}{c}{Performances}  \\ 
 & InRes (H×W) & PatchSz (H×W) & FeatDim & Params (B) & FLOPs (G) & VQA\textsuperscript{v2} & GQA & VQA\textsuperscript{T} & POPE & MME \\ \hline
Siglip & $384 \times 384$ & $14 \times 14$ & $728 \times 1152$ & 0.428 & 670.89 & 75.5 & 59.2 & 44.1 & 85.7 & 1179.0 \\ 
TinyCLIP-39M & $224 \times 224$ & $16 \times 16$ & $196 \times 512$ & 0.038 & 15.04 & 67.1 & 52.9 & 31.1 & 79.2 & 1069.5 \\
TinyCLIP-40M & $224 \times 224$ & $32 \times 32$ & $49 \times 512$ & 0.039 & 3.93 & 64.2 & 51.9 & 30.8 & 77.9 & 1022.3 \\
TinyCLIP-61M & $224 \times 224$ & $32 \times 32$ & $49 \times 640$ & 0.061 & 6.10 & 63.7 & 51.5 & 34.4 & 77.6 & 1010.2 \\ \hline
\end{tabular}
}
\end{table*}

We apply HiMix across several popular language decoders to assess its generalizability, with Siglip as the vision encoder in all experiments.

\noindent \textbf{Regular Paradigm Results.}
The results, as shown in Table~\ref{tab-main_2Stage}, indicate that HiMix maintains comparable performance to the baseline model on most benchmarks, while significantly reducing computational cost. This demonstrates that our method achieves a successful balance between efficiency and effectiveness. We noticed a decline in model performance on TextVQA and MME tasks, primarily due to a decrease in OCR capability.

% We observe some accuracy declines in TextVQA and MME tasks under the regular training setup. Our analysis of MME revealed significant differences in OCR scores across various subcategories compared to the original model. Notably, the TextVQA paper also suggests that OCR can improve accuracy in such tasks. The relative disadvantage of HiMix in these tasks may be due to the introduced vision projection layers requiring more thorough training, which could enhance its capabilities in fine-grained vision understanding.

\noindent \textbf{Enhanced Paradigm Results.}
To achieve better performance, we evaluated HiMix using the Enhanced Paradigm strategy. Table~\ref{tab-main_3Stage} shows that adding additional training data improves performance. Figure~\ref{fig-gap} illustrates performance gaps between the HiMix and the baseline model under two training strategies, demonstrating that Enhanced Paradigm amplifies advantage of HiMix, generally outperforms the baseline in this setup.

% \begin{figure*}[htb]
%     \centering
%     \begin{minipage}[t]{0.65\linewidth}
%         \centering
%         \includegraphics[width=\linewidth,height=5cm]{figure/fig-gap.pdf}
%         \caption{Performance gaps between the HiMix and the baseline model across various benchmarks under two training strategies: \textcolor[HTML]{F0906D}{base data} and \textcolor[HTML]{46756B}{expanded data}. The y-axis shows the performance gap relative to the baseline, with a positive value indicating an improvement. In (a), the baseline is LLaMA3-1B, while in (b), the baseline is Qwen2-0.5B. The results demonstrate that expanding the training data amplifies the performance benefits of HiMix.}
%         \label{fig-gap}
%     \end{minipage}
%     \hfill
%     \begin{minipage}[t]{0.32\linewidth}
%         \centering
%         \includegraphics[width=\linewidth,height=5cm]{figure/fig-VE.pdf}
%         \caption{Comparison of different vision encoders. All models use TinyLlama as the language decoder. The y-axis represents the model's performance on $VQA^{v2}$, while the x-axis indicates the computational cost of the vision encoder. The size of each circle is proportional to the model's parameter count.}
%         \label{fig-ve}
%     \end{minipage}
% \end{figure*}

\noindent \textbf{Efficiency Analysis.}
We conduct a comprehensive analysis of the computational efficiency of HiMix compared to baseline LVLM models. The evaluation includes the computational cost (GFLOPs) of the language decoder and the maximum VRAM usage across varying ratios of vision-to-language input lengths. As shown in Table~\ref{tab-effi}, models using HiMix demonstrate substantial reductions in computational costs and memory usage. Specifically, scenarios with V:L of 728:32 and 728:64 closely resemble typical multimodal input configurations. HiMix achieves a significant reduction in computational cost when language sequences are much shorter than vision sequences. Even as the length of the language sequence increases, HiMix consistently exhibits lower computational costs and VRAM consumption. When the input vision and language sequence lengths are equal, HiMix consumes only half the computational cost of baseline models. This suggests that HiMix is well-suited for both general applications and more complex language processing tasks, providing substantial efficiency advantages over traditional methods.

% \noindent \textbf{Efficiency Analysis.}
% We conduct a comprehensive analysis of the computational efficiency of HiMix compared to baseline LVLM models. The evaluation includes Time-to-First-Token (TTFT), the computational cost (GFLOPs) of the language decoder, and the maximum VRAM usage across varying ratios of vision-to-language input lengths. As shown in Table~\ref{tab-effi}, models using HiMix demonstrate substantial reductions in computational costs and memory usage, along with improvements in response time. Specifically, scenarios with V:L of 728:32 and 728:64   closely resemble typical multimodal input configurations. HiMix achieves a significant reduction in computational cost when language sequences are much shorter than vision sequences. Even as the length of the language sequence increases, HiMix consistently exhibits lower computational costs and VRAM consumption. When the input vision and language sequence lengths are equal, HiMix consumes only half the computational cost of baseline models. This suggests that HiMix is well-suited for both general applications and more complex language processing tasks, providing substantial efficiency advantages over traditional methods.

\subsection{Extended Evaluation}
By adopting HiMix, the computational cost of the language decoder significantly decreases, with the vision encoder accounts for the majority of the overall computational cost. To further reduce this cost, we explore the use of smaller vision encoders to evaluate model performance under minimized computational demands. Specifically, we compare the performance of TinyCLIP series~\cite{wu2023tinyclip} and Siglip as vision encoders, using TinyLlama-1.1B as the language decoder. The results are shown in Table ~\ref{tab-abla_ve}. 

For TinyCLIP-39M and TinyCLIP-40M, While these two encoders have the same feature dimensions, TinyCLIP-39M has a smaller patch size, enabling finer-grained image processing. Results show that TinyCLIP-39M outperformed across metrics.
For TinyCLIP-40M and TinyCLIP-61M, Although these encoders have the same token count,  TinyCLIP-61M features a larger dimensionality. However, increasing feature dimensions did not enhance performance. Instead, patch size, which affects feature sequence length, exerts a more substantial impact.

While using a smaller vision encoder reduces overall performance, it also achieves a substantial reduction in computational cost. In scenarios with stringent constraints on computational resources, such as deployment on edge/mobile devices, HiMix with a smaller vision encoder can greatly facilitate the feasibility of deploying LVLMs.

Some additional experimental results and visualizations have been provided in the \textcolor{blue}{Appendix}.

\section{Discussion and conclusion}
%\noindent %\textbf{Limitations.} Due to constraints in computational resources, our experiments were conducted on smaller-scale LLMs. We intend to expand our evaluations to larger models to provide a more comprehensive view of HiMix.

%讨论：计算量的减少意味着推理速度的提升，这在一些需要实时计算分析的场景是十分重要的。同时意味着计算资源消耗的减少，这在一些需要考虑设备电量消耗的场景里是十分珍贵的。
\noindent 
\textbf{Discussion.} The reduction in computational workload implies an improvement in inference speed and a decrease in computational resource consumption, which is extremely valuable in scenarios that require real-time computing analysis or need to consider device power consumption.
Due to constraints in computational resources, our experiments were conducted on smaller-scale LLMs. We intend to expand our evaluations to larger models to provide a more comprehensive view of HiMix.

\noindent \textbf{Conclusion.}  The conventional information interaction in LVLMs suffers from high computational costs, 
caused by vision sequences in LLM module computation. To address this issue, this paper introduces a novel mechanism for vision-language interaction, proposing a Mixture Attention module that restructures the composition and interaction of the LLM input sequence, significantly reducing the overall computational load in LVLMs. Furthermore, we develop a hierarchical injection mixture attention architecture to improve the model performance by integrating diverse vision information at specific stages within the LLM. Extensive experimental results across multiple LVLMs validate the effectiveness of our approach. Our method not only reduces computational complexity substantially but also maintains performance comparable to that of original LVLMs, showing strong potential to meet a wider range of application requirements.
{
    \small
    \bibliographystyle{ieeenat_fullname}
    \bibliography{main}
}

\clearpage
\setcounter{page}{1}
\maketitlesupplementary
\appendix
\renewcommand{\thesection}{\Alph{section}}
In the supplementary materials, the following sections are included:

\begin{itemize}
    
\item  \textbf{Implementation Details in Section~\ref{section A}.} 

This section provides a comprehensive description of the experimental setup, including detailed parameter settings. The aim is to ensure the reproducibility and clarity of the experiments.
\item  \textbf{Additional Experimental Results in Section~\ref{section B}.} 

This section consists of two parts: a detailed comparison of other methods and HiMix, and an in-depth breakdown of HiMix’s computational costs.
\item  \textbf{Qualitative Results in Section~\ref{section C}.} 

This section showcases detailed visualizations of the performance of HiMix across various tasks.

\end{itemize}

\section{Implementation Details}
\label{section A}
Table~\ref{params} summarizes the detailed hyperparameters used during training under the Regular Paradigm. For the Enhanced Paradigm, the hyperparameters for the Further Pretraining and Finetuning stages are identical to those in the Finetuning stage of the Regular Paradigm. All experiments are conducted on NVIDIA A100 GPUs.

\setlength{\tabcolsep}{10pt}
\begin{table}[h]
\caption{Hyperparameter Settings for Pretrain and Finetune}
\label{params}
\resizebox{\linewidth}{!}{
\begin{tabular}{ccc}
 \hline   
Parameter                                                                & Pretrain                                               & Finetune                                                   \\  \hline   
Training Modules & \begin{tabular}[c]{@{}c@{}}Vision \\ Proj\end{tabular} & \begin{tabular}[c]{@{}c@{}}Language\\ Decoder\end{tabular} \\
Learning Rate    & 1e-3                                                   & 2e-5                                                       \\
Batch Size       & 256                                                    & 128                                                        \\
LR Schedule      & \multicolumn{2}{c}{Cosine decay}                                                                                    \\
Optimizer        & \multicolumn{2}{c}{AdamW}                                                                                           \\
Weight Decay     & \multicolumn{2}{c}{0}                                                                                               \\
Zero Stage       & \multicolumn{2}{c}{Zero 3}                                                                                          \\
Warmup Ratio     & \multicolumn{2}{c}{0.03}                                                                                            \\
Data Precision   & \multicolumn{2}{c}{bf16}                                                                                            \\
Attention        & \multicolumn{2}{c}{Flash Attention 2}          \\ \hline                                                                    
\end{tabular}
}
\end{table}

As projection layers for processing vision information are newly added to each layer of the Language Decoder and are randomly initialized, the number of training epochs is adjusted for different models to ensure sufficient training. Specifically:
\begin{itemize}
   
\item[-] For the Qwen~\cite{yang2024qwen2} and Llama~\cite{dubey2024llama} series, the pretraining stage is conducted over three epochs.
\item[-] For the TinyLlama~\cite{zhang2024tinyllama}, the pretraining stage is conducted over ten epochs.
\item[-] For all models, the Further Pretraining stage is conducted over one epoch, and the Finetuning stage is conducted over three epochs.
\end{itemize}
% Figure~\ref{loss} shows the training loss of HiMix based on four language models under the Regular Paradigm.

% \begin{figure}[h]
%     \centering
% \includegraphics[width=\linewidth]{figure/loss.pdf}
%     \caption{Training Loss Curves of HiMix Across Four Language Models Under the Regular Paradigm}
%     \label{loss}
% \end{figure}

\section{Additional Results} \label{section B}
\subsection{Complementary Experiment on 7B Model}
\begin{table*}[!htp]
\caption{Comprehensive comparison of performance and computational efficiency between the original model and HiMix, using SigLIP as the vision encoder. Performance metrics include VQAv2, GQA, TextVQA, MM-Vet, POPE, MME, and MMMU. Computational efficiency is assessed by Language Decoder parameter count (B) and GFLOPs. Performance improvements are highlighted in \textbf{bold}, with computational cost shown in \textbf{\textcolor[HTML]{FF0000}{red}} as a percentage of the original model.}
\label{tab-7b}
\makebox[\textwidth][c]{ 
\resizebox{\textwidth}{!}{
\begin{tabular}{cccccccccc}
\hline
                           & \multicolumn{7}{c}{Performances}                                                                                                                                                                                                                                                                                                                                                                                                                                                                           & \multicolumn{2}{c}{Efficiency}                                                                                                 \\
\multirow{-2}{*}{\begin{tabular}[c]{@{}c@{}}Language\\ Decoder\end{tabular}}& \multicolumn{1}{c}{VQA\textsuperscript{v2}~\cite{goyal2017vqav2}}                               & \multicolumn{1}{c}{GQA~\cite{hudson2019gqa}}                                  & \multicolumn{1}{c}{VQA\textsuperscript{T}~\cite{singh2019textvqa}}                     & \multicolumn{1}{c}{MM-Vet~\cite{Yu2023MMVetEL}}                               & \multicolumn{1}{c}{POPE~\cite{li2023pope}}                        & \multicolumn{1}{c}{MME~\cite{Fu2023MMEAC}}                           & MMMU~\cite{yue2024mmmu}                                & \multicolumn{1}{c}{\begin{tabular}[c]{@{}c@{}}Params\\ (B)\end{tabular}} & \begin{tabular}[c]{@{}c@{}}FLOPs\\ (G)\end{tabular} \\ \hline
Vicuna-7B                                         & \multicolumn{1}{c}{80.6}                                 & \multicolumn{1}{c}{63.3}                                 & \multicolumn{1}{c}{63.7}                        & \multicolumn{1}{c}{40.0}                                 & \multicolumn{1}{c}{86.6}                        & \multicolumn{1}{c}{1439.1}                        & 36.4                                & \multicolumn{1}{c}{{\color[HTML]{333333} 7.19}}                          & 10800                                                 \\ 
+HiMix                                             & \multicolumn{1}{c}{{80.3}} & \multicolumn{1}{c}{{62.4}} & \multicolumn{1}{c}{61.4}                        & \multicolumn{1}{c}{{37.5}} & \multicolumn{1}{c}{\textbf{87.4}}                        & \multicolumn{1}{c}{\textbf{1498.9}}                        & {\textbf{38.3}} & \multicolumn{1}{c}{7.47}                                                 & { \textbf{1310\color[HTML]{FE0000}(12\%)}}             \\ \hline

\end{tabular}
}
}
\end{table*}
HiMix employs vision projection layers to obtain richer visual features, resulting in an increase in model parameters compared to the original model. According to the Scaling Law, model performance typically improves with an increase in the number of parameters and the amount of training data. If model parameters are increased without a corresponding increase in training data, the model may not fully realize its potential. Therefore, to achieve optimal performance with HiMix, we augment the training data. The training data is sourced from LLaVA-OneVision~\cite{Li2024LLaVAOneVisionEV}, with approximately 4M data used during the pretrain stage (from LLaVA-OneVision Stage 1.5) and about 3.2M data during the finetuning stage(from LLaVA-OneVision Stage 2-SI). The visual encoder is Siglip, and the language decoder is Vicuna-7b. Both the original model and the HiMix are trained for one epoch in each of the two stages, with results presented in Table ~\ref{tab-7b}.
Compared to the original model, HiMix achieves similar performance while reducing the computational load of the decoder to only 12\%. Both Table~\ref{tab-7b} and the experimental results in the main paper demonstrate that HiMix can significantly reduce computational overhead while maintaining performance, regardless of whether applied to small-scale or large-scale MLLMs.

\begin{table*}[!h]
\caption{Performance and computational efficiency comparison of Llama3-1B, FastV (under different configurations with K for filtering layer and R for filtering ratio), and HiMix. FLOPs are computed under an input configuration of V:L=728:64.}
\label{cam-fastv}
\resizebox{\linewidth}{!}{
\begin{tabular}{lcccccccc}
\hline
Language Decoder    & VQA\textsuperscript{v2}~\cite{goyal2017vqav2}                     & GQA~\cite{hudson2019gqa}                        & VQA\textsuperscript{T}~\cite{singh2019textvqa}                    & MM-Vet~\cite{Yu2023MMVetEL}                      & POPE~\cite{li2023pope}                        & MMMU~\cite{yue2024mmmu}                        & \begin{tabular}[c]{@{}c@{}}FLOPs\\ (G)\end{tabular} & \begin{tabular}[c]{@{}c@{}}FLOPs\\ Ratio\end{tabular} \\ \hline
Llama3-1B           & {\color[HTML]{1F2328} 76.8} & {\color[HTML]{1F2328} 59.6} & {\color[HTML]{1F2328} 52.7} & {\color[HTML]{1F2328} 27.8} & {\color[HTML]{1F2328} 86.7} & {\color[HTML]{1F2328} 30.6} & {\color[HTML]{1F2328} 2040}                         & 100\%                                                 \\
+FastV (K=2 R=90\%) & 62.9                        & 51.0                        & 42.6                        & 17.4                        & 71.9                        & 30.0                        & 510                                                 & 25\%                                                  \\
+FastV (K=2 R=75\%) & 72.2                        & 56.3                        & 48.4                        & 24.5                        & 84.1                        & 29.8                        & 758                                                 & 37\%                                                  \\
+FastV (K=2 R=50\%) & 75.4                        & 58.9                        & 50.7                        & 26.5                        & 87.6                        & 30.8                        & 1180                                                & 58\%                                                  \\
+FastV (K=3 R=90\%) & 60.5                        & 49.2                        & 38.8                        & 17.7                        & 69.6                        & 30.2                        & 595                                                 & 29\%                                                  \\
+FastV (K=3 R=75\%) & 70.8                        & 55.1                        & 45.5                        & 24.0                        & 82.8                        & 30.3                        & 829                                                 & 41\%                                                  \\
+FastV (K=3 R=50\%) & 75.0                        & 58.5                        & 49.7                        & 27.7                        & 87.0                        & 30.8                        & 1230                                                & 60\%                                                  \\
+FastV (K=5 R=90\%) & 65.8                        & 51.5                        & 39.9                        & 18.3                        & 73.6                        & 30.0                        & 765                                                 & 38\%                                                  \\
+FastV (K=5 R=50\%) & 75.6                        & 59.0                        & 50.3                        & 28.6                        & 87.2                        & 30.8                        & 1320                                                & 65\%                                                  \\

\rowcolor[HTML]{D3D3D3} +HiMix     & 76.2                        & 59.6                        & 50.5                        & 28.4                        & 86.6                        & 29.9                        & \textbf{192}                                        & \textbf{9\% }     \\ \hline                                            
\end{tabular}
}
\end{table*}

\begin{table*}[!h]
\caption{Component-wise Computational Costs (FLOPs) of the Language Decoder Across Different V:L Ratios. This table provides a detailed breakdown of the FLOPs consumed by attention (denoted as F\_Attn) and feed-forward networks (denoted as F\_FFN) within the language decoder for different vision-to-language input length ratios.}
\label{com-cost}
\resizebox{\linewidth}{!}{
\begin{tabular}{ccccccccccc}
\hline
\multirow{2}{*}{Method} & \multicolumn{2}{c}{V:L 728:32} & \multicolumn{2}{c}{V:L 728:64} & \multicolumn{2}{c}{V:L 728:200} & \multicolumn{2}{c}{V:L 728:728} & \multicolumn{2}{c}{V:L 728:1000} \\
                        & F\_Attn        & F\_FFN        & F\_Attn        & F\_FFN        & F\_Attn         & F\_FFN        & F\_Attn         & F\_FFN        & F\_Attn         & F\_FFN                            \\ \hline
Qwen2-0.5B              & 117                                 & 476                                & 124                                 & 497                                & 156                                  & 582                                & 311                                  & 914                                & 410                                  & 1085                                \\
\rowcolor[HTML]{D3D3D3}+HiMix                  & 15                                  & 19                                 & 20                                  & 40                                 & 44                                   & 126                                & 166                                  & 457                                & 248                                  & 628                                 \\
TinyLlama-1.1B          & 420                                 & 1157                               & 442                                 & 1206                               & 541                                  & 1413                               & 988                                  & 2217                               & 1258                                 & 2631                                \\
\rowcolor[HTML]{D3D3D3}+HiMix                  & 37                                  & 49                                 & 55                                  & 97                                 & 136                                  & 304                                & 513                                  & 1109                               & 747                                  & 1523                                \\
Llama-3.2-1B            & 331                                 & 1224                               & 348                                 & 1276                               & 425                                  & 1495                               & 768                                  & 2345                               & 973                                  & 2783                                \\
\rowcolor[HTML]{D3D3D3}+HiMix                  & 41                                  & 50                                 & 56                                  & 103                                & 119                                  & 322                                & 411                                  & 1173                               & 590                                  & 1611                                \\
Llama-3.2-3B            & 1270           & 3213          & 1333           & 3349          & 1605            & 3924          & 2783            & 6156          & 3465            & 7306           \\
\rowcolor[HTML]{D3D3D3}+HiMix                  & 148            & 131           & 204            & 270           & 442             & 846           & 1488            & 3078          & 2101      & 4228      \\ \hline       
\end{tabular}
}
\end{table*}

\subsection{Comparison with Other Methods}
In the Related Work section of the main paper, we introduced several approaches to Vision Token Reduction. To comprehensively evaluate our proposed method, we select one of the most competitive methods, FastV (ECCV 2024 Oral)~\cite{chen2024image}, and compare it with our proposed method, HiMix. FastV introduces a dynamic image tokens pruning method to reduce the inference cost of LVLMs. In contrast, HiMix avoids the forward propagation of the entire vision sequence in the language decoder, instead leveraging a mixture attention mechanism to interact with the language at specific stages within each layer.

As shown in Table~\ref{cam-fastv}, we present a detailed comparison of FastV and HiMix in terms of both model performance and computational efficiency. Using Llama3-1B as the baseline model, we evaluate FastV under various parameter configurations and compare it against HiMix. While FastV achieves a reduction in computational cost compared to the baseline, its overall performance declines significantly as the computational costs decrease (i.e., as more vision tokens are pruned). Even under its best-performing configuration (Filtering layer K=5, Vision token filtering ratio R=50\%), FLOPs of FastV remain at 65\% of the original model.

In sharp contrast, HiMix maintains performance comparable to the baseline while reducing FLOPs to just 9\% of the original model. This represents a striking improvement in computational efficiency. HiMix demonstrates clear advantages in both model performance and efficiency, highlighting its practicality and strong potential for real-world applications where both accuracy and computational efficiency are essential.

\subsection{Detailed Computational Analysis}
To provide a more in-depth analysis of the efficiency results presented in the main paper, this subsection examines the computational cost breakdown within the language decoder under different input configurations. Specifically, Table~\ref{com-cost} presents the FLOPs (in GFLOPs) consumed by the two major components of the language decoder—attention and feed-forward networks (FFN). The V:L ratio represents the lengths of the vision and language sequences in the input prompt. Computational costs are reported for both the baseline models and the proposed HiMix.

As described in Equation 7 of the main paper, HiMix avoids the quadratic complexity of Vanilla-LVLM in the attention mechanism. For the FFN, the vision sequence is excluded entirely from computation, which significantly reduces the overall computational cost.

The results highlight HiMix's ability to optimize the computational cost of the language decoder while preserving performance metrics, positioning it as a practical solution for deploying LVLMs in resource-constrained scenarios with a strong balance between accuracy and efficiency.

The tool used for testing FLOPs is ~\cite{calflops}.

\section{Qualitative Results} \label{section C}
\begin{figure*}[h]
    \centering
    \includegraphics[width=\linewidth]{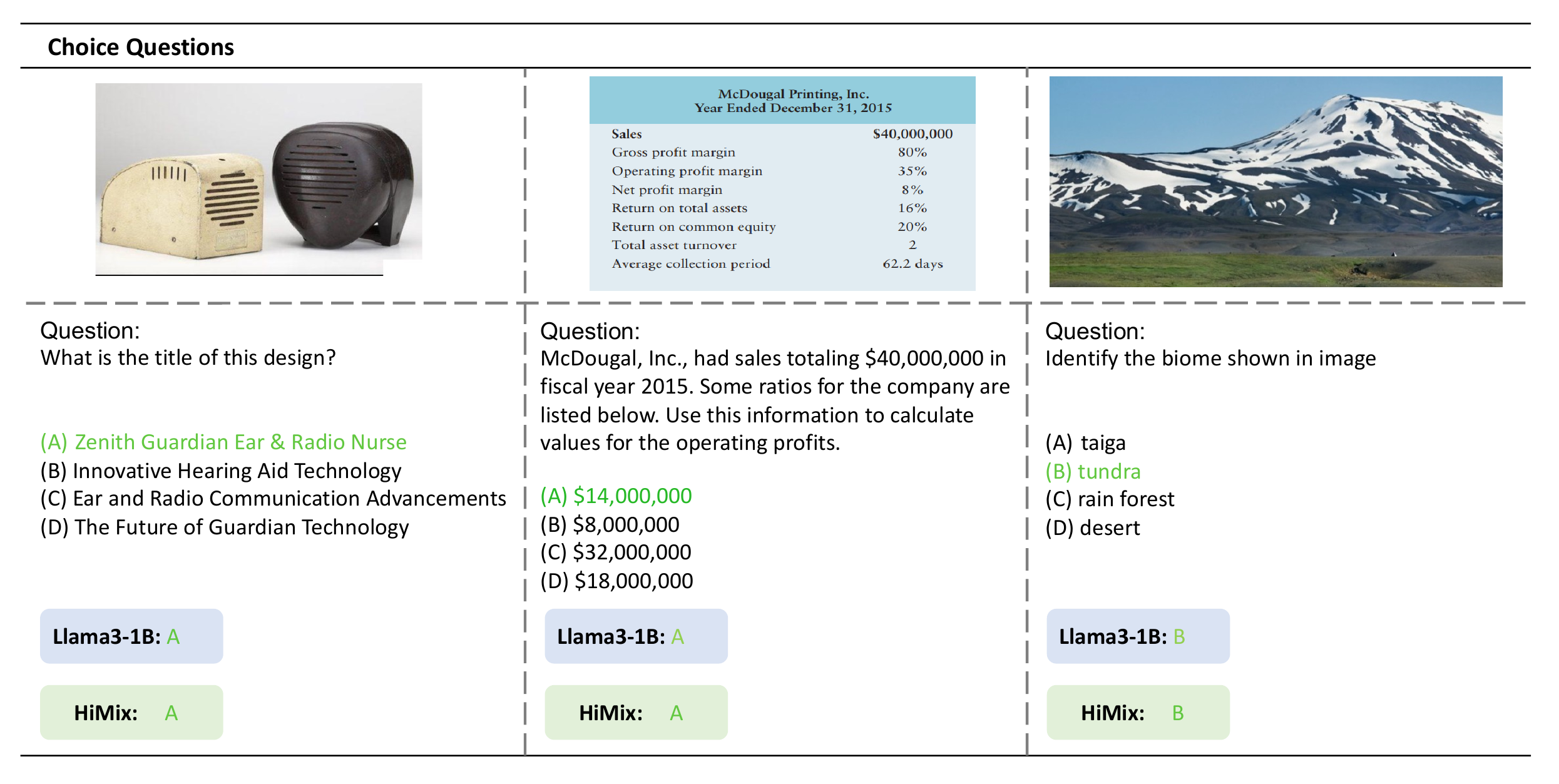}
    \caption{Qualitative results of choice questions.}
    \label{sup_choice}
\end{figure*}

\begin{figure*}[h]
    \centering
    \includegraphics[width=\linewidth]{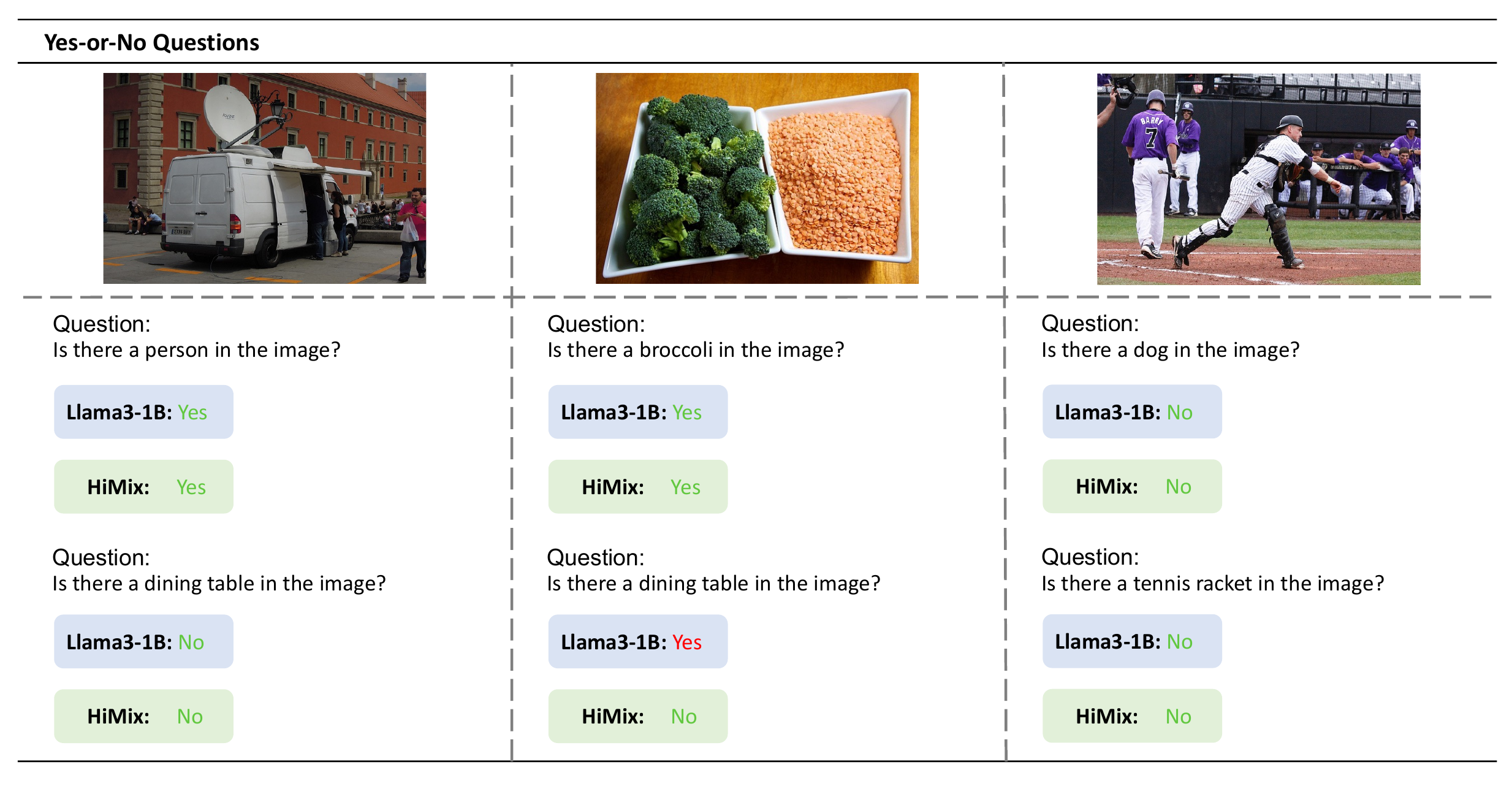}
    \caption{Qualitative results of yes/no questions. {\color[HTML]{62C840}Correct answers} and {\color[HTML]{FF0000}Wrong answers} are highlighted in color respectively.}
    \label{sup_yesno}
\end{figure*}

\begin{figure*}[h]
    \centering
    \includegraphics[width=\linewidth]{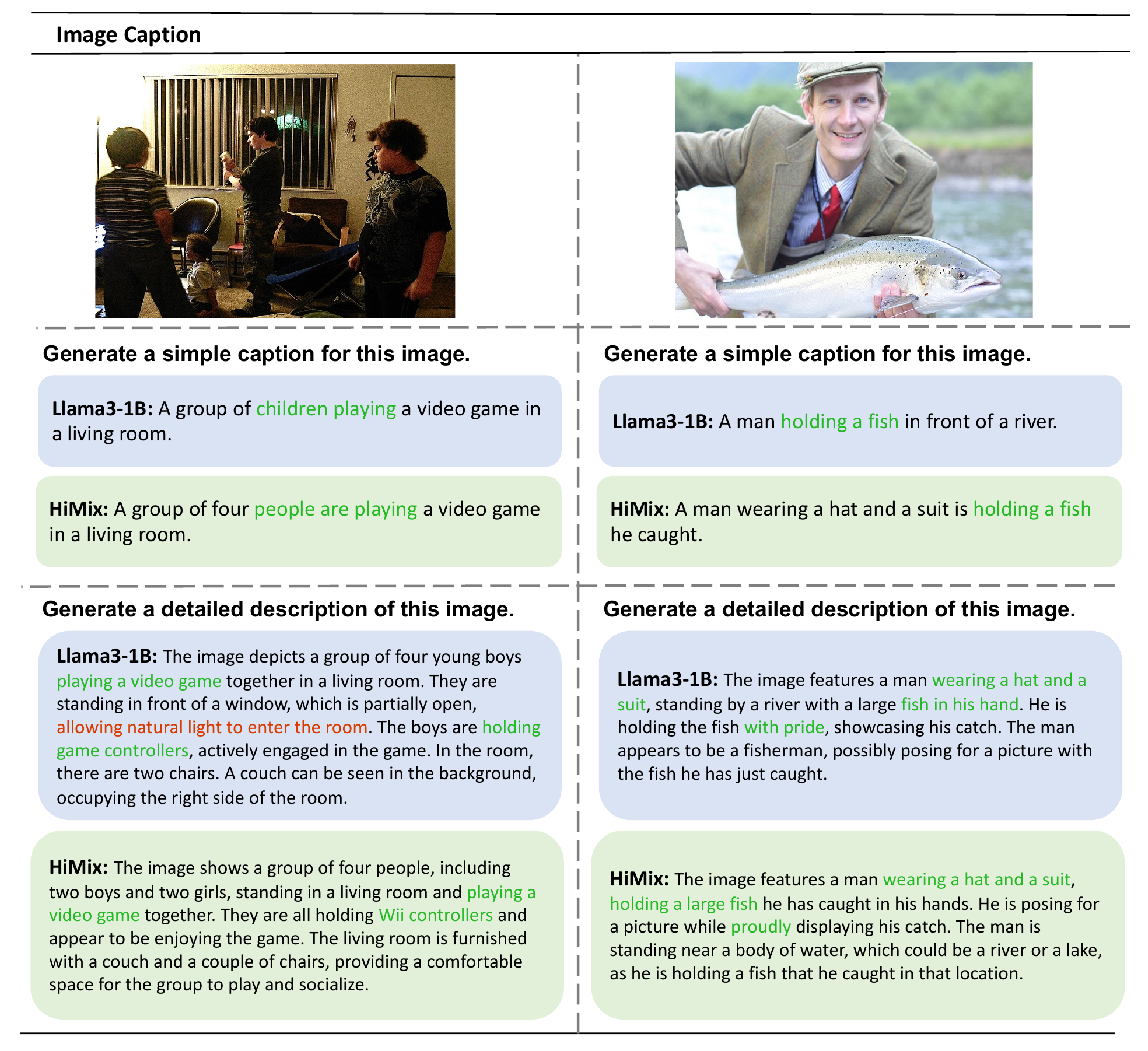}
    \caption{Qualitative comparison of image captions. {\color[HTML]{62C840}Correct descriptions} and {\color[HTML]{FF0000}hallucinated} are highlighted in color respectively.}
    \label{sup_ic1}
\end{figure*}

\begin{figure*}[h]
    \centering
    \includegraphics[width=\linewidth]{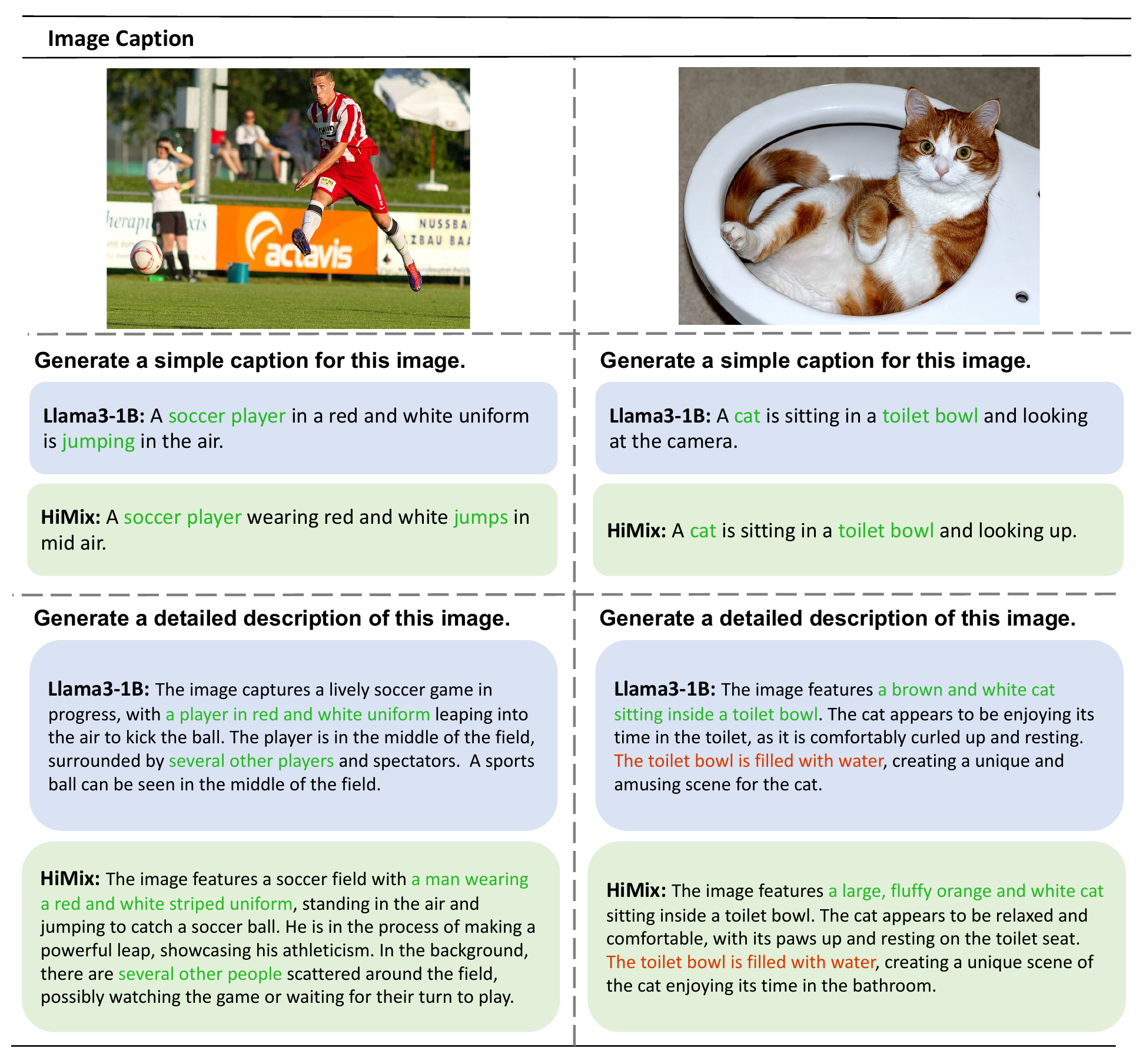}
    \caption{Qualitative comparison of image captions. {\color[HTML]{62C840}Correct descriptions} and {\color[HTML]{FF0000}hallucinated} are highlighted in color respectively.}
    \label{sup_ic2}
\end{figure*}

\begin{figure*}[h]
    \centering
    \includegraphics[width=\linewidth]{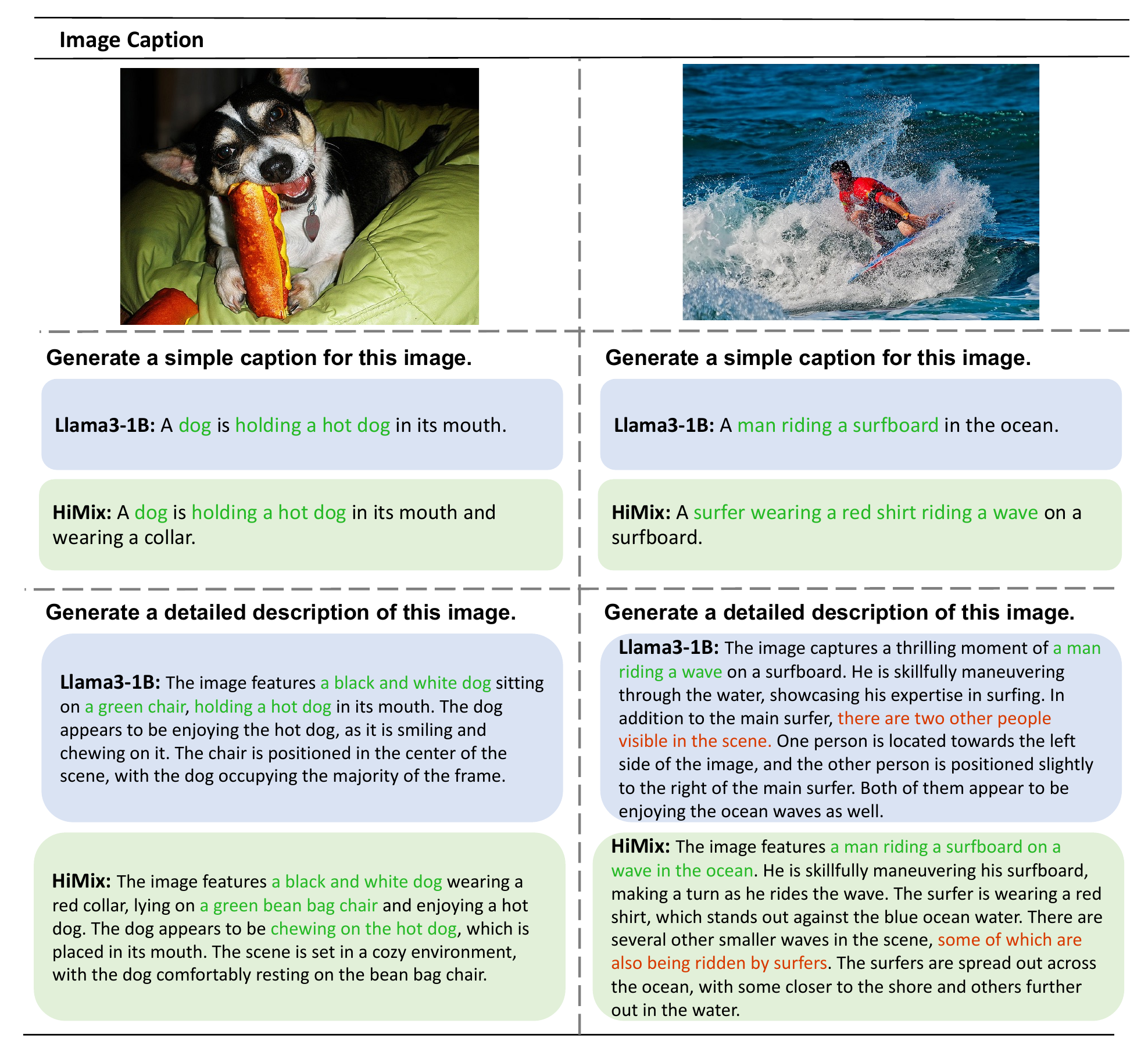}
    \caption{Qualitative comparison of image captions. {\color[HTML]{62C840}Correct descriptions} and {\color[HTML]{FF0000}hallucinated} are highlighted in color respectively.}
    \label{sup_ic3}
\end{figure*}

\begin{figure*}[h]
    \centering
    \includegraphics[width=0.95\linewidth]{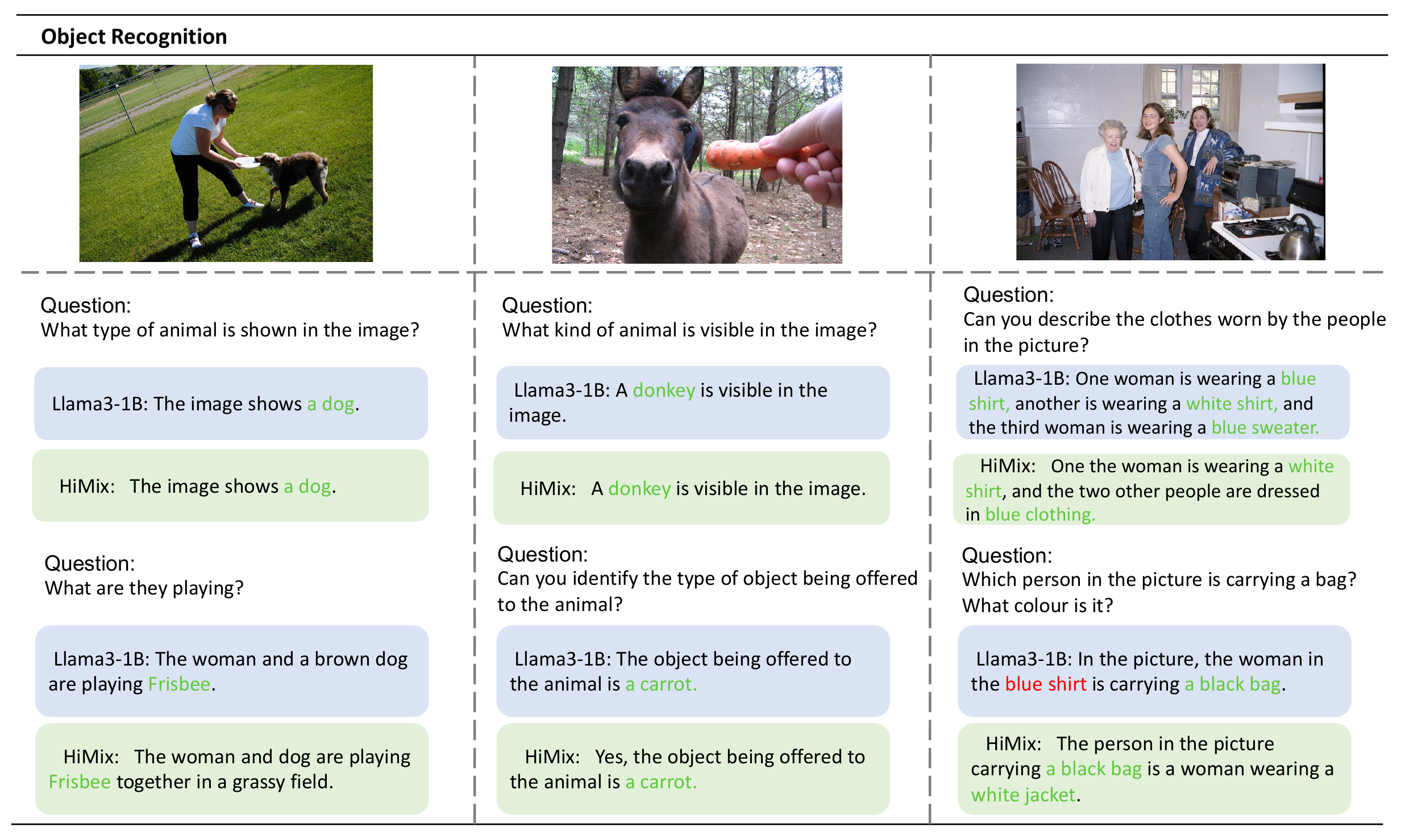}
    \caption{Qualitative results of object recognition tasks. {\color[HTML]{62C840}Correctly identified objects} and {\color[HTML]{FF0000}errors} are highlighted in color respectively.}
    \label{sup_obj}
\end{figure*}

\begin{figure*}[ht]
    \centering
    \includegraphics[width=0.95\linewidth]{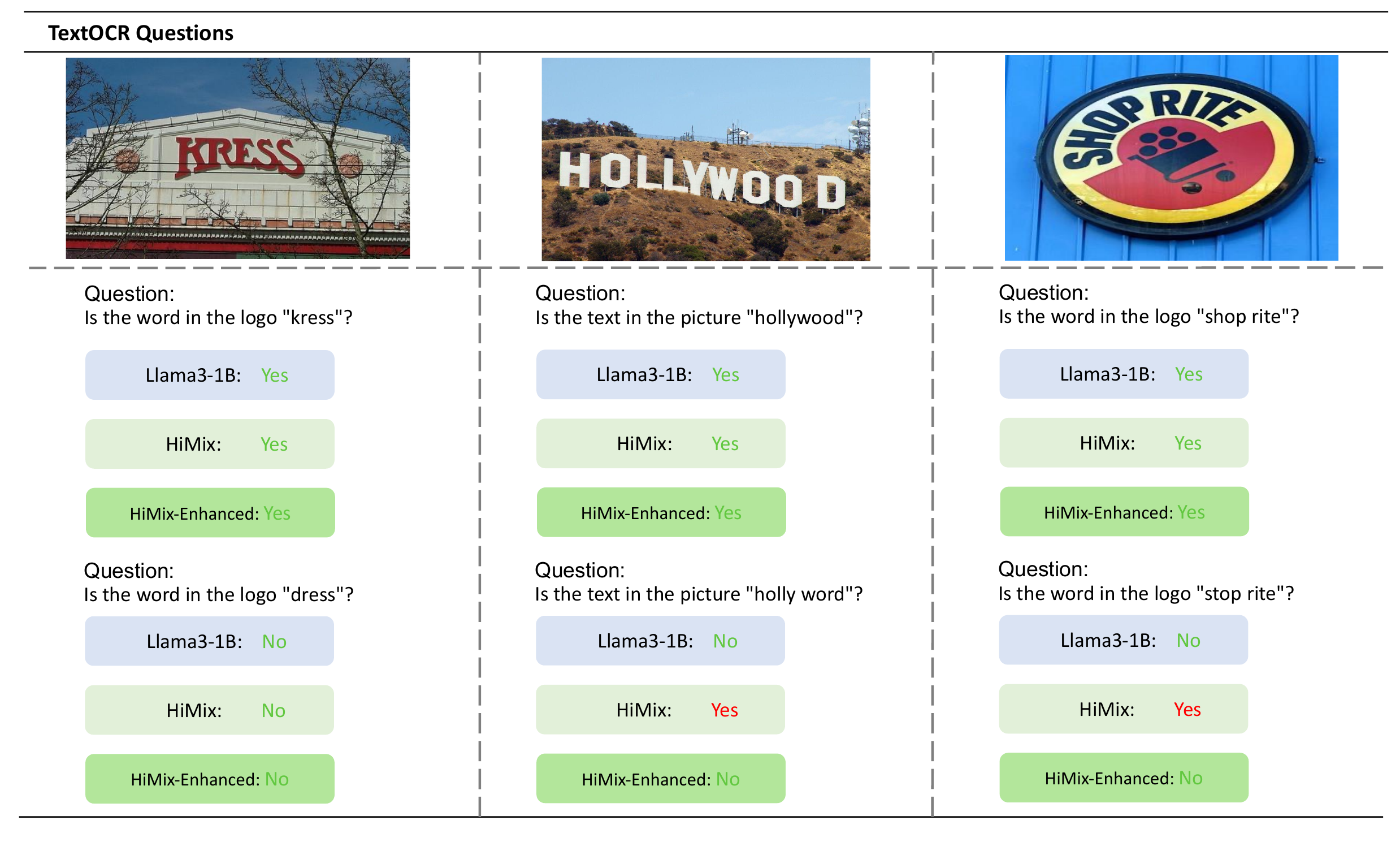}
    \caption{Qualitative results of text OCR tasks. {\color[HTML]{62C840}Correctly recognized text} and {\color[HTML]{FF0000}errors} are highlighted for comparison.}
    \label{sup_textocr}
\end{figure*}

This section provides qualitative results to offer a more intuitive understanding and comparison between the baseline model (e.g., Llama3-1B) and HiMix. Specifically, we showcase multiple tasks, including choice questions (Figure~\ref{sup_choice}), yes/no questions (Figure~\ref{sup_yesno}), simple image captions and detailed image captions (Figures~\ref{sup_ic1}, ~\ref{sup_ic2}, and ~\ref{sup_ic3}), object recognition (Figure~\ref{sup_obj}) and text OCR (Figure~\ref{sup_textocr}). We observe the following results:
\begin{itemize}
    
\item HiMix achieves performance comparable to the baseline. Across various multimodal tasks, HiMix accurately understands and responds to the requirements, demonstrating strong alignment with task-specific objectives.

\item When the task requires detailed image descriptions, both the baseline model and HiMix occasionally generate similar hallucinated descriptions (highlighted in red in the images). We attribute this behavior to the inherent limitations of the selected baseline model. Nevertheless, these results confirm that HiMix does not compromise the baseline model's ability to process and interpret vision information, maintaining performance parity with the baseline model.

\item As discussed in the main paper, HiMix exhibits a slight decline in OCR capabilities. We present examples of failure cases to illustrate this limitation. However, under the Enhanced Paradigm, this drawback is mitigated, demonstrating that training data augmentation effectively improves performance.

\end{itemize}

\end{document}